%% file: main.tex
\documentclass[10pt,twocolumn,letterpaper]{article}

\usepackage{iccv}
\input{preamble}

\input{defs}

\usepackage[pagebackref=true,breaklinks=true,letterpaper=true,colorlinks,bookmarks=false]{hyperref}

\iccvfinalcopy

\begin{document}

\title{BabelCalib: A Universal Approach to Calibrating Central Cameras}

\makeatletter
\renewcommand*{\@fnsymbol}[1]{\ifcase#1\or\textdagger\else\@arabic{#1}\fi}
\makeatother
\author{
Yaroslava Lochman\textsuperscript{1,}\thanks{Part of this work was done when Yaroslava Lochman and James Pritts were at Facebook Reality Labs.} \ \ \ \ \
Kostiantyn Liepieshov\textsuperscript{3} \ \ \ \ \
Jianhui Chen\textsuperscript{2} \\[3pt]
Michal Perdoch\textsuperscript{2} \ \ \ \ \ \
Christopher Zach\textsuperscript{1} \ \ \ \ \ \ 
James Pritts\textsuperscript{1,}$^\dagger$ \\[3pt]
\small
\textsuperscript{1}Chalmers University of Technology \ \ \ \ \
\textsuperscript{2}Facebook Reality Labs \ \ \ \ \ \textsuperscript{3}Ukrainian Catholic University \\
{\tt\footnotesize \{lochman, zach, pritts\}@chalmers.se} \ \ \ \
{\tt\footnotesize liepieshov@ucu.edu.ua} \ \ \ \
{\tt\footnotesize \{jchen2020, mperdoch\}@fb.com}
\vspace{-10pt}
}

\maketitle

\begin{abstract}
  Existing calibration methods occasionally fail for
  large field-of-view cameras due to the non-linearity of the
  underlying problem and the lack of good initial values for all
  parameters of the used camera model. This might occur because a
  simpler projection model is assumed in an initial step, or a poor
  initial guess for the internal parameters is pre-defined.  A lot of
  the difficulties of general camera calibration lie in the use of a
  forward projection model. We side-step these challenges by first
  proposing a solver to calibrate the parameters in terms of a
  back-projection model and then regress the parameters for a target
  forward model. These steps are incorporated in a robust estimation
  framework to cope with outlying detections.  Extensive experiments
  demonstrate that our approach is very reliable and returns the most
  accurate calibration parameters as measured on the downstream task
  of absolute pose estimation on test sets.  The code is released at
  \href{https://github.com/ylochman/babelcalib}{\texttt{https://github.com/ylochman/babelcalib}}.
\end{abstract}

\input{introduction}
\input{preliminaries}
\input{initial_guess}
\input{method}

\input{experiments}
\input{conclusions}

{\small
\bibliographystyle{ieee_fullname}
\bibliography{main}
}

\newpage
\newpage
\input{supplemental}

\end{document}

%% file: preamble.tex
\usepackage{times}
\usepackage{epsfig}
\usepackage{graphicx}
\usepackage{amsmath}
\usepackage{amssymb}
\usepackage{amsfonts}
\usepackage{bm}
\usepackage{booktabs}
\usepackage[scale=2]{ccicons}
\usepackage[thinc]{esdiff}
\usepackage{xspace}
\usepackage{mathtools}
\usepackage[inline,shortlabels]{enumitem}
\usepackage{makecell}
\usepackage{multirow}
\usepackage[ruled,vlined]{algorithm2e}
\usepackage{color}
\usepackage{xcolor}

\usepackage{placeins}

\usepackage{caption} 

%% file: defs.tex
\newcommand{\sksym}[1]{\ensuremath{\left[{#1}\right]_{\times}}\xspace}

\newcommand{\ma}[1]{\ensuremath{\mathtt{#1}}\xspace}
\def\m#1{\ensuremath{\mathtt{#1}}\xspace}

\def\mA{{\m A}}

\def\mF{{\m F}}

\def\mP{{\m P}}

\def\mR{{\m R}}
\def\mK{{\m K}}

\def\mT{{\m T}}

\def\mH{{\m H}}

\def\vh{\ensuremath{\mathbf{h}}}

\def\ve{\ensuremath{\mathbf{e}}}
\def\vr{\ensuremath{\mathbf{r}}}

\def\vu{\ensuremath{\mathbf{u}}}

\def\vt{\ensuremath{\mathbf{t}}}

\def\vzero{\ensuremath{\mathbf{0}}}

\def\deg{^\circ}
\def\tr{^\top}

\def\inv#1{#1^\mathsf{-1}}
\def\opt#1{#1^*}

\providecommand{\homogvec}[1]{\binom{n}{1}}

\DeclareMathOperator{\atantwo}{atan2}

\newcommand{\cspond}[2]{\ensuremath{#1\leftrightarrow#2}}

\makeatletter
\def\munderbar#1{\underline{\sbox\tw@{$#1$}\dp\tw@\z@\box\tw@}}
\makeatother

\newcommand{\vX}[1][]{\ensuremath{\mathbf{X}_{#1}}\xspace}

\newcommand{\vx}[1][]{\ensuremath{\mathbf{x}_{#1}}\xspace}
\newcommand{\vxp}[1][]{\ensuremath{\vx[#1]^{\prime}}\xspace}

\newcommand{\secref}[1]{Sec.~{\ref{#1}}}
\newcommand{\figref}[1]{Fig.~{\ref{#1}}}
\newcommand{\tabref}[1]{Table~{\ref{#1}}}

\newcommand{\tabsref}[1]{Tables~{\ref{#1}}}
\newcommand{\algref}[1]{Algorithm~{\ref{#1}}}

\newcommand{\R}[1][]{\ensuremath{\mathbb{R}^{#1}}\xspace}

\DeclarePairedDelimiter{\diagfences}{(}{)}
\newcommand{\diag}{\operatorname{diag}\diagfences}

\newcommand{\T}{{\!\top}}

\newcommand{\buildset}[3]{\ensuremath{ \{\,#1\,\}_{#2}^{#3} }\xspace}
\newcount\colveccount
\newcommand*\colvec[1]{
    \global\colveccount#1
    \begin{pmatrix}
    \colvecnext
}
\def\colvecnext#1{
    #1
    \global\advance\colveccount-1
    \ifnum\colveccount>0
        \\
        \expandafter\colvecnext
    \else
        \end{pmatrix}
    \fi
}
\newtoks\rowvectoks
\newcommand{\rowvec}[2]{
  \rowvectoks={#2,}\count255=#1\relax
  \advance\count255 by -1
  \rowvecnexta}
\newcommand{\rowvecnexta}{
  \ifnum\count255>0
    \expandafter\rowvecnextb
  \else
    \setlength\arraycolsep{1pt}     
    \begin{pmatrix}\the\rowvectoks\end{pmatrix}
  \fi}
\newcommand\rowvecnextb[1]{
  \ifnum\count255>1     
    \rowvectoks=\expandafter{\the\rowvectoks&#1,}
  \else
    \rowvectoks=\expandafter{\the\rowvectoks&#1}
  \fi
    \advance\count255 by -1
    \rowvecnexta}

\makeatletter
\DeclareRobustCommand\onedot{\futurelet\@let@token\@onedot}
\def\@onedot{\ifx\@let@token.\else.\null\fi\xspace}
\def\eg{\emph{e.g}\onedot} \def\Eg{\emph{E.g}\onedot}

\def\etal{\emph{et al}\onedot}
\makeatother

\DeclareMathOperator*{\argmin}{argmin}

\newcommand{\RANSAC}{RANSAC\xspace}

\makeatletter

\def\addlegendimage{\pgfplots@addlegendimage}
\makeatother

\definecolor{mycolor1}{rgb}{0,0,0}
\definecolor{mycolor2}{rgb}{1,0,1}
\definecolor{mycolor3}{rgb}{0,1,1}
\definecolor{mycolor4}{rgb}{0,0,1}
\definecolor{mycolor5}{rgb}{0,1,0}
\definecolor{betteryellow}{rgb}{1,0.8824,0.0980}
\definecolor{lavender}{rgb}{0.902,0.7451,1.0}
\definecolor{olive}{rgb}{0.5020,0.5020,0} 
\definecolor{orange}{rgb}{1,0.5,0} 
\definecolor{bettergreen}{rgb}{0,0.6,0.3}

\definecolor{blue}{HTML}{4069B0}
\definecolor{lightorange}{HTML}{FF8F00}
\definecolor{orange}{HTML}{E45611}
\definecolor{darkorange}{HTML}{B85325}
\definecolor{lightgreen}{HTML}{34DC5B}
\definecolor{green}{HTML}{28A745}
\definecolor{lightgray}{HTML}{656972}
\definecolor{gray}{HTML}{53585F}
\definecolor{darkgray}{HTML}{333333}
\definecolor{red}{HTML}{D61901}
\definecolor{magenta}{HTML}{CC00CC}
\definecolor{cyan}{HTML}{00FFFF}

\def\fb#1{\textcolor{bettergreen}{\textbf{#1}}}
\def\sb#1{\textcolor{lightorange}{\textbf{#1}}}
\def\ffb#1{\textbf{#1}}
\def\red#1{\textcolor{red}{\textbf{#1}}}
\def\green#1{\textcolor{green}{\textbf{#1}}}

\def\opencvBC{{OpenCV-BC}\xspace}
\def\opencvKB{{OpenCV-KB}\xspace}
\def\opencvUCM{{OpenCV-UCM}\xspace}

\def\kalibrBC{{Kalibr-BC}\xspace}
\def\kalibrKB{{Kalibr-KB}\xspace}
\def\kalibrUCM{{Kalibr-UCM}\xspace}
\def\kalibrEUCM{{Kalibr-EUCM}\xspace}
\def\kalibrFOV{{Kalibr-FOV}\xspace}
\def\kalibrDS{{Kalibr-DS}\xspace}

\def\occDiv{{OCamCalib-DIV}\xspace}

\def\ourBC{\textbf{Ours-BC}\xspace}
\def\ourKB{\textbf{Ours-KB}\xspace}
\def\ourUCM{\textbf{Ours-UCM}\xspace}
\def\ourEUCM{\textbf{Ours-EUCM}\xspace}
\def\ourFOV{\textbf{Ours-FOV}\xspace}
\def\ourDS{\textbf{Ours-DS}\xspace}
\def\ourDiv{\textbf{Ours-DIV}\xspace}

\def\ovcorner{\texttt{OV-Corner}\xspace}
\def\ovcube{\texttt{OV-Cube}\xspace}
\def\ovplane{\texttt{OV-Plane}\xspace}
\def\kalibrdata{\texttt{Kalibr}\xspace}
\def\occdata{\texttt{OCamCalib}\xspace}
\def\uzhdavis{\texttt{UZH-DAVIS}\xspace}
\def\uzhsnap{\texttt{UZH-Snapdragon}\xspace}

%% file: introduction.tex
\renewcommand{\arraystretch}{1.6}
\begin{table*}[!t]
  \begin{center}
  \footnotesize
  \begin{tabular*}{\textwidth}{cl@{\extracolsep{\fill}}cc}
    \toprule
    & \textsc{Model} & \textsc{Parameters, $\theta$} & \textsc{Radial (Back-)Projection Function} \\
    \midrule\vspace{2pt}
    \multirow{6}{*}{\rotatebox{90}{\hspace*{-16pt} \emph{\scriptsize{Forward}}}}
    & Brown-Conrady (BC)~\cite{Brown-P-1971}
    & {$\{k_1, k_2\}$}
    & {$\phi_\theta(R,Z) = (R/Z) \cdot \left(1 + \sum_{n=1}^2 k_n \left(R/Z \right)^{2n}\right)$} \\[2pt]
    & Kannala-Brandt (KB)~\cite{Brandt-PAMI-2006}
    & {$\{k_1,\dotsc,k_4\}$}
    & {$\phi_\theta(R,Z) = \zeta + \sum_{n=1}^4 k_n \zeta^{2n+1}, \quad \zeta = \atantwo(R,Z)$} \\[2pt]
    & Unified Camera (UCM)~\cite{Mei-ICRA-2007}
    & {$\{\xi\}$}
    & {$\phi_\theta(R,Z) = R (\xi+1) / (\xi (\sqrt{R^2+Z^2}) + Z)$} \\[2pt]
    & Field of View (FOV)~\cite{Devernay-MVA01}
    & {$\{w\}$}
    & {$\phi_\theta(R,Z) = \frac{1}{w} \atantwo{(2R \tan\frac{w}{2},Z)}$} \\[2pt]
    & Extended Unified Camera (EUCM)~\cite{Khomutenko-RAL15}
    & {$\{\alpha, \beta\}$}
    & {$\phi_\theta(R,Z) = R/(\alpha d + (1 - \alpha)Z)$}, \quad $d=\sqrt{\beta R^2+Z^2}$ \\[2pt]
    & Double Sphere (DS)~\cite{Usenko-3dv-2020}
    & {$\{\xi,\alpha\}$}
    & {$\phi_\theta(R,Z) = R/(\alpha d_2 + (1-\alpha)Z_2), \quad d_2 = \sqrt{R^2+Z_2^2}, \,\, Z_2=\xi \sqrt{R^2+Z^2} + Z$} \\[2pt]
    \midrule
    \multirow{2}{*}{\rotatebox{90}{\hspace*{-5pt}\emph{\scriptsize{Backward}}}}
    & Division (DIV)~\cite{Scaramuzza-ICVS-2006,Urban-ISPRS-2015}
    & {$\{a_1,a_2,a_3\}$}
    & $\psi_\theta(r) = 1 + \sum_{n=1}^3 a_n r^{n+1}$ \\[2pt]
    & Division-Even~\cite{Larsson-ICCV19}
    & $\{\lambda_1,\dotsc,\lambda_N\}$
    & $\psi_\theta(r) = 1 + \sum_{n=1}^N \lambda_n r^{2n}$\\
    \bottomrule
  \end{tabular*}
  \caption{\textbf{Supported camera models.} Models compute
    either radially-symmetric projection, $\,\, r = \phi_\theta(R,Z)$,
    or back-projection, $\,\, r:\,\, r Z - R \psi_\theta(r) = 0$,
    where $R$ and $Z$ are the radial and depth components of a scene
    point, and $r$ is the distance from the center of projection of a
    retinal point. The right column lists functions for published
    models.
  }
  \label{tab:models}
  \end{center}
  \vspace{-20pt}
  \end{table*}
  \renewcommand{\arraystretch}{1}

\section{Introduction}
Cameras with very wide fields of view, such as fisheye lenses and
catadioptric rigs~\cite{Zhang-ICRA16}, usually require highly
nonlinear models with many parameters.  Calibrating these cameras can
be a tedious process because of the camera model's complexity and its
underlying non-linearity. If the calibration is inaccurate or even
fails, then the user is often required to manually remove problematic
images or fiducials, capture additional images, or provide better
initial guesses for the unknown model parameters. A second common
problem is that the choice of calibration toolbox limits the user to a
particular set of supported camera models and extending the toolbox to
accommodate more flexible camera models can be a difficult task.

This paper proposes a method that robustly estimates accurate camera
models for central projection cameras~\cite{Sturm-BOOK11} with fields
of view ranging from both narrow to omni-directional. Furthermore, the
proposed framework can estimate the most widely used camera models for
these lens types, while also providing an easy and common path to
extend the method to new camera models.

\begin{figure}
  \begin{center}
    \begin{tabular*}{0.478\textwidth}{cc}
    \hspace*{-10pt}
    \includegraphics[height=0.248\textwidth]{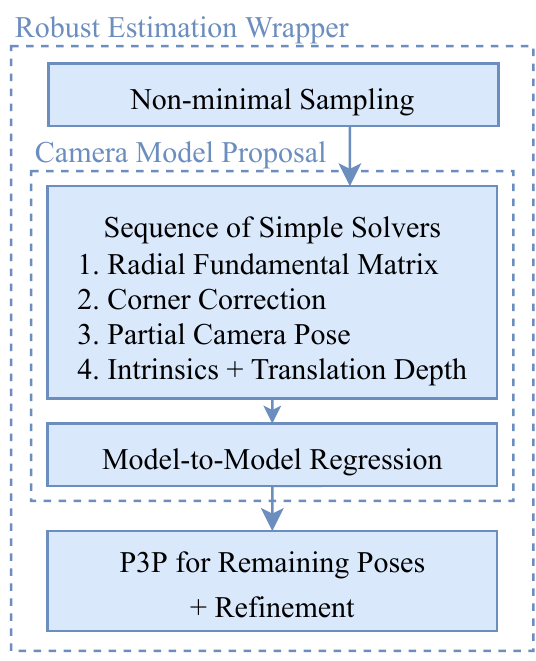}
    &
    \hspace*{-13pt}
    \includegraphics[height=0.248\textwidth]{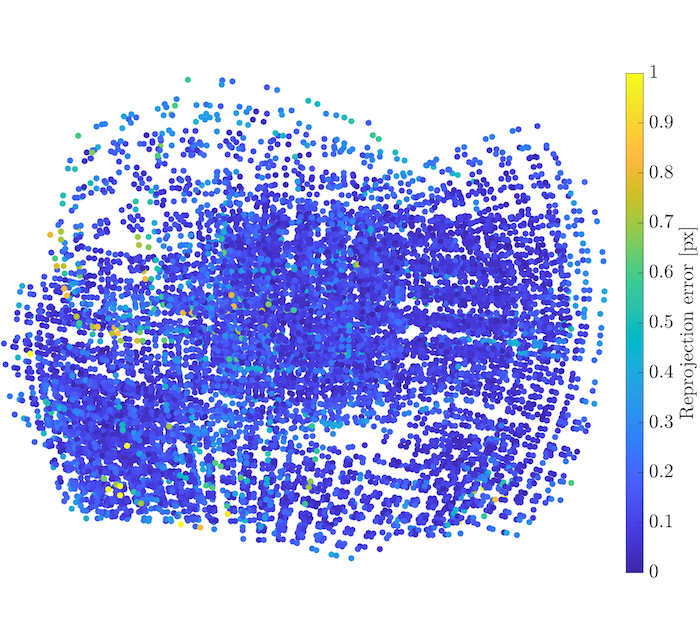}
    \end{tabular*}
    \vspace{-6pt}
    \caption{\textbf{Method overview and result.} (left) BabelCalib
      pipeline: the camera model proposal step ensures a good
      initialization, (right) example result showing residuals of
      reprojected corners of test images.}
  \label{fig:teaser}
  \end{center}
\vspace{-25pt}
\end{figure}

Camera calibration is a very non-linear task, hence a good initial
guess is typically needed to obtain accurate parameters. Poor initial
estimates are frequent source of failures. Sensible initial guesses
are often available only for some of the model unknowns, \eg, initial
values are often unavailable for parameters describing substantial
lens distortions.

A second failure mode is caused by incorrect or grossly inaccurate
measurements, \eg, corner detections, which are matched to fiducials
on the calibration target. If corrupted data is used to estimate the
initial guess, then the downstream model refinement will likely fail.

Our method addresses both failure modes. We introduce a solver that
recovers all calibration parameters for a wide range of cameras (and
lenses) such as pinhole, fisheye and catadioptric ones.  We show that
the proposed solver provides a good initialization for all critical
intrinsics, which includes the center of projection and pixel aspect
ratio. Our solver assumes only a planar calibration target. In
addition, the initialization simultaneously improves the accuracy of
corner detections while estimating the center of projection and camera
pose by enforcing projective constraints. The solver is used within a
\RANSAC framework for efficient model generation. The model proposals
are evaluated for consistency with the extracted features, and poorly
extracted features and incorrect correspondences are rejected.

Our approach uses a back-projection model as an intermediate camera
model.  Back-projection models mapping image points to 3D ray
directions are able to model a wide range of cameras (such as pinhole,
fisheye and omni-directional cameras). Our approach (and therefore our
main contribution) is to decouple the calibration task for general
camera models into a much simpler calibration task for a powerful
back-projection model followed by a regression task to obtain the
parameters of the general target camera model. Effectively, we remove
the need to generate a solver for each target camera model, which can
be intractable, or result in solvers that are computationally
expensive or numerically unstable in practice. Instead, we use an
efficient solver for a back-projection model followed by an easier
regression task to recover the target model parameters. The motivation
for such an approach is given in Table~\ref{tab:models}, where it
shows that projection equations are relatively simple once the radial
component $R$ and the depth component $Z$ are known. These values are
provided by the back-projection model. \Eg, for the Kannala-Brandt
model~\cite{Brandt-PAMI-2006}, estimation of its parameters is linear
for given $R$ and $Z$.

Overall, our stratified approach to calibration circumvents many of
the issues that are due to the highly non-linear behavior of many
flexible camera models. BabelCalib performs both back-projection
estimation and target model regression inside of a robust estimation
framework, and directly returns the parameters of the target
model. \figref{fig:teaser} illustrates the accuracy of our method for
a fisheye lens on hold-out test images. The achieved high accuracy is
spatially coherent across the entire calibration target over all test
images.

\subsection{Related Work}
\label{sec:related_work}

Camera calibration is an important tool in order to upgrade cameras
from pure imaging devices to geometric sensors, and it has led to the
development of many parametric models for cameras (and their lens
systems) and the introduction of respective toolboxes
(\eg~\cite{Brown-P-1971,heikkila1997four,Zhang-PAMI00,Fitzgibbon-CVPR01,Sturm-ECCV04,claus2005rational,Mei-ICRA-2007,Li-ICIRS-2013,Ramalingam-PAMI16}).
In order to facilitate the highest accuracy for the calibration
parameters, a controlled, usually planar calibration target is
employed in many applications. The use of dedicated images (``training
data'') for the task of camera calibration distinguishes standard
calibration from self-calibration, that extracts calibration
parameters from uncontrolled ``test'' images
(\eg~\cite{faugeras1992camera,hartley1994self,pollefeys1999self,fraser1997digital,Wildenauer-CVPR12,Pritts-PAMI20,Lochman-WACV21}).

Most relevant to our approach in terms of forward projection models are the
unified camera model~\cite{geyer2000unifying,barreto2001issues,Mei-ICRA-2007},
the fisheye projection model by Kannala and Brandt~\cite{Brandt-PAMI-2006},
and the double sphere model~\cite{Usenko-3dv-2020}. Using these models
for calibration tasks is not always straightforward and comes  with their own
set of assumptions.

\Eg, the estimator proposed for the Double-Sphere model~\cite{Usenko-3dv-2020}
requires that the circular field-of-view is visible to recover the center of
projection and the aspect ratio, that the relative position of the spherical
retinas be initialized, and that a non-radial line be identified to recover
the focal length.
Further, the method proposed
for the popular Kannala-Brandt model requires specifying focal length and
field of view~\cite{Brandt-PAMI-2006}.

The introduction of a linear solver to calibrate the division model in
the back-projection framework~\cite{Scaramuzza-ICVS-2006} demonstrates
the benefits of using back-projections. This linear method assumes
known center of distortion and unit aspect ratio and is extended to a
two-stage method in~\cite{Scaramuzza-IROS-2006} to include estimation
of the center of distortion. Urban \etal \cite{Urban-ISPRS-2015}
identifies the shortcomings of this two-stage method and suggests
joint refinement of all unknowns instead.

Finally, very general non-parametric models for cameras and lenses have been
proposed
(including~\cite{ramalingam2005towards,Hartley-PAMI-2007,Camposeco-ICCV15,Schops-CVPR20}),
Our experiments indicate that appropriate parametric
models are sufficient to model a wide range of cameras and lenses and are
therefore---due to Occam's razor---preferable in general.

%% file: preliminaries.tex
\section{Preliminaries}
\label{sec:preliminaries}
Let us define the camera matrix $\mP$ mapping from scene
coordinates to ray directions in the camera coordinate system as $\mP
= \diag{f,f,1}\begin{bmatrix} \mR & \vt \end{bmatrix}$, where $f$ is the focal
length, $\mR = \begin{bmatrix} \vr_1 & \vr_2 & \vr_3 \end{bmatrix}$ is the rotation matrix,
and $\vt = \rowvec{3}{t_x}{t_y}{t_z}^{\T}$ is the translation vector.
We build on the omni-directional camera model of Micusik and Pajdla~\cite{Micusik-ICCV-2003},
that relates the image point $\vu=\rowvec{3}{u}{v}{1}^{\T}$ and the scene point $\vX$ as
\begin{equation}
  \label{eq:omni_camera_model}
  \gamma g(\ma{A}\vu) = \mP\vX \quad \text{s.t.} \quad \gamma > 0.
\end{equation}
In \eqref{eq:omni_camera_model} the matrix $\ma{A}$ maps from image
coordinates to sensor coordinates. Denote the center of projection as
$\ve=\rowvec{3}{e_x}{e_y}{1}^{\T}$, the scale factor as $s$, and the
pixel aspect ratio as $a$. For the initialization method we assume
that the pixels are orthogonal, so we have
\begin{equation}
  \ma{A} = \diag{\inv{a},1,1}\diag{\inv{s},\inv{s},1} \ma{T}(-\ve),
\end{equation}
where
$\ma{T}(\vx)$ is a homogeneous matrix encoding translation by \vx.
The nonlinear function $g(\cdot)\in\R[3]\to\R[3]$ in
\eqref{eq:omni_camera_model} maps from the retinal plane to ray
directions in the camera coordinate system. For the initialization
method, the typically small distortions caused by lens
misalignment are ignored~\cite{Hartley-BOOK04} so that we can model
back-projection of $\vu=\rowvec{3}{u}{v}{1}^{\T}$ as radially
symmetric, $g(\vu) = \rowvec{3}{u}{v}{\psi(r(\vu))}^{\T}$, where the
radius of the retinal point is $r(\vu)=\sqrt{u^2+v^2}$.

\paragraph{Back-Projection with Division Model}
We parameterize $\psi(\cdot)$ with the division model. It has a good ability
to model significant lens distortions and was
used in \cite{Scaramuzza-ICVS-2006} for fisheye and
catadioptric lenses with fields-of-view greater than $180^\circ$.
The model is defined as
\begin{equation}
  \label{eq:bproj_division_model}
  \psi(r) = 1 + \sum_{n=1}^N \lambda_n r^{2n}
\end{equation}
Function $\psi(\cdot)$ is not invertible in general; however, we
assume that there is only a single root $\opt{r}\in [0, r^{\max}]$, where $r^{\max}$ is the image diagonal. More
precisely, let $\vx=\rowvec{3}{x}{y}{w}^{\T}$ and let $h(\cdot)$ be the function projecting $\vx$ to
the retinal plane,
\begin{align}
  \label{eq:proj_division_model}
    h(\vx) = \rowvec{3}{\frac{\opt{r}}{r(\vx)}x}{\frac{\opt{r}}{r(\vx)}y}{1}^\T,
\end{align}
where $\opt{r}$ is the only solution of $\psi(r)=w$ in $[0,r^{\max}]$. 
Consequently,
\begin{equation}
  \label{eq:invertible}
  \vu = h(g(\vu)).
\end{equation}
Multiple roots in $[0,r^{\max}]$ imply that a scene point maps to multiple
image points, which is an implausible physical configuration.

\paragraph{Radial Fundamental Matrix}
Without loss of generality, we assume the target to be on the plane
$z=0$. Transforming a point \vX on the target to a ray direction in
the camera coordinate system can be done by the homography $\ma{H}
= \begin{bmatrix} \vr_1 & \vr_2 & \vt \end{bmatrix}$ constructed from
the camera matrix
\begin{equation}
  \label{eq:camera_viewing_scene_plane}
  \begin{split}
    \mP\vX &= \diag{f,f,1}\begin{bmatrix} \mR & \vt \end{bmatrix} \rowvec{4}{X}{Y}{0}{1}^{\T} = \\
    & \diag{f,f,1} \underbrace{\begin{bmatrix}\vr_1 & \vr_2 &
        \vt \end{bmatrix}}_{\ma{H}}\underbrace{\rowvec{3}{X}{Y}{1}^{\T}}_{\vx}. \\
  \end{split}
\end{equation}
Hartley and Kang \cite{Hartley-PAMI-2007} used the radial fundamental
matrix to recover the principal point of distorted pinhole cameras. We
extend the radial fundamental matrix to recover the center of
projection $\ve$ and camera pose $\mR, \vt$ for the back-projection model of  \cite{Micusik-ICCV-2003}.
We substitute $\diag{f,f,1}\ma{H}\vx$ for \mP\vX in
\eqref{eq:omni_camera_model} using
\eqref{eq:camera_viewing_scene_plane}, apply the projection function
$h(\cdot)$ to both sides, and use \eqref{eq:invertible} to eliminate
$g$ giving
\begin{equation}
  \label{eq:radial_scaling0}
  \begin{split}
    \vu = \inv{\ma{A}}h(1/\gamma\diag{f,f,1}\ma{H}\vx) = \\
    \mT(\ve)1/\gamma\diag{a(sfr^*)/r(\vx),(sfr^*)/r(\vx),1}\ma{H}\vx.
  \end{split}
\end{equation}
Note that the scaling by $(sfr^*)/(\gamma r(\vx))$ due to focal length,
projection by $h(\cdot)$, pixel scale $s$ and depth multiplier $\gamma$ is entangled and acts radially.
Substituting $\eta=(sfr^*)/(\gamma r(\vx))$ into \eqref{eq:radial_scaling0} and crossing both sides with \ve\xspace gives
\begin{equation}
  \label{eq:radial_scaling3}
  \begin{split}
  \sksym{\ve} \vu &= \sksym{\ve} \ma{T}(\ve)\diag{a\eta,\eta,1/\gamma}\ma{H}\vx \\
  &= \sksym{\ve}\diag{a\eta,\eta,0}\ma{H}\vx.
  \end{split}
\end{equation}

The radial line $\sksym{\ve}\vu$ is eliminated by taking the
inner product of $\vu\xspace$ with \eqref{eq:radial_scaling3}, and
$\eta$ can be eliminated since it is non-zero. Denote the rows of
$\ma{H}$ such that $\ma{H}=\begin{bmatrix} \vh_1 & \vh_2 &
\vh_3 \end{bmatrix}\tr$. Then \eqref{eq:radial_scaling3} simplifies to
\begin{equation*}
0 = \vu\tr \sksym{\ve}\rowvec{3}{a \vh_1^{\T}\vx}{\vh_2^{\T}\vx}{0}\tr,
\end{equation*}
which can be rearranged to give the radial fundamental matrix
$\ma{F}_r$ for omni-directional cameras
\begin{equation}
  0 = \vu\tr \underbrace{\sksym{\ve}
  \begin{bmatrix}a r_{11} & a r_{12} & a t_x \\
    r_{21} & r_{22} & t_y \\
    0 & 0 & 0
  \end{bmatrix}}_{\ma{F}_r} \vx.
  \label{eq:radial_fundamental_matrix} 
\end{equation}
The aspect ratio is modeled, but cannot be recovered without
additional constraints.  The radial fundamental matrix $\ma{F}_r$ is
rank two by construction, and the center of projection \ve\xspace is a
basis for the left null space of $\ma{F}_r$.

%% file: initial_guess.tex
\section{Obtaining the Initial Estimate}
\label{sec:good_guess}
The methods proposed in this section ensure that a good initial guess
of the camera model is made. The parameters are recovered by a
sequence of simple solvers (see \figref{fig:teaser}). The
back-projection model of \eqref{eq:omni_camera_model} corresponds
image points to ray directions in the camera coordinate system. Given
this correspondence, we show that regressing the commonly used
projection models can be done easily.
This enables the search for
good initial guesses for the target projection model in a sampling
framework, which increases the robustness of the method.

\subsection{Solving the Radial Fundamental Matrix}
\label{sec:solve_F}
The radial fundamental matrix is estimated to recover the center of
projection and camera pose. The epipolar constraint on the radial
fundamental matrix $\vu\tr \ma{F} \vx=0$ in
\eqref{eq:radial_fundamental_matrix} can be written as a linear
constraint on $\ma{F}$
\begin{equation}
  \label{eq:billinear_form}
  (\vx \otimes \vu)^\T \operatorname{vec}(\ma{F}) = 0.
\end{equation}
Following the classic solver for the fundamental matrix in
\cite{Hartley-BOOK04}, we use at least seven image-to-target point
correspondences, denoted $\{\,\cspond{\vu_i}{\vx[i]}\,\}$, to compute
the null space of stacked constraints of the form
\eqref{eq:billinear_form}. The nonlinear constraint $\det \ma{F} = 0$
is enforced to recover at most three real solutions from the null
space.  The fundamental matrix consistent with the most
correspondences is kept.

\subsection{Solving the Center of Projection and Pose}
\label{sec:solving_center_of_projection}
As shown in \eqref{eq:radial_fundamental_matrix}, the center of
projection is a basis for the left null space of $\ma{F}$
\begin{equation}
  \label{eq:principal_point}
  \zeta \ve = \operatorname{null}{\ma{F}} \tr.
\end{equation}
There is a scale ambiguity, denote it $\nu$, between the radial
fundamental matrix $\ma{F}_r$ as formulated in
\eqref{eq:radial_fundamental_matrix} and the fundamental matrix
$\ma{F}$ recovered by the seven-point method,
\begin{equation}
\ma{F} = \nu \ma{F}_r \quad \text{where } \ma{F} = \begin{pmatrix} f_{ij} \end{pmatrix}.
\end{equation}

Let $r_{31},r_{32}$ be the unknown components of the rotation vectors
$\vr_1$ and $\vr_2$.  If we let $\ma{S}=\nu^{-1}\diag{{a}^{-1},1,1}$,
then $\vr_j = \ma{S} \rowvec{3}{f_{2j}}{-f_{1j}}{r_{3j}}^\T$. We use
the orthonormality of $\vr_1$ and $\vr_2$ to put quadratic constraints
on the unknowns,
\begin{equation}
  \begin{split}
  &\|\ma{S} (f_{21},-f_{11},r_{31})\tr\|^2_2 = \|\ma{S} (f_{22},-f_{12}, r_{32})\tr\|^2_2 \\
  & \text{ and } \rowvec{3}{f_{21}}{-f_{11}}{r_{31}} \ma{S}^2 \rowvec{3}{f_{22}}{-f_{12}}{r_{32}}\tr = 0.
  \end{split}
  \label{eq:orthonormality_constraints}
\end{equation}
There are four unknowns but only three constraint
equations. Additional constraints are needed to recover the aspect
ratio. The unknowns $\{\nu,
a,r_{31},r_{32},t_z,\lambda_1,...,\lambda_n\}$ can be jointly
recovered by solving a system of polynomial equations (see
\secref{sec:supp_aspect_ratio_solver} in Supplemental); however, we chose to sample over the interval of aspect
ratios $a \in [0.5,2]$ and recover $\{\nu, r_{31}, r_{32}\}$ from
\eqref{eq:orthonormality_constraints}.

\begin{figure}
\begin{center}
\includegraphics[width=0.47\textwidth]{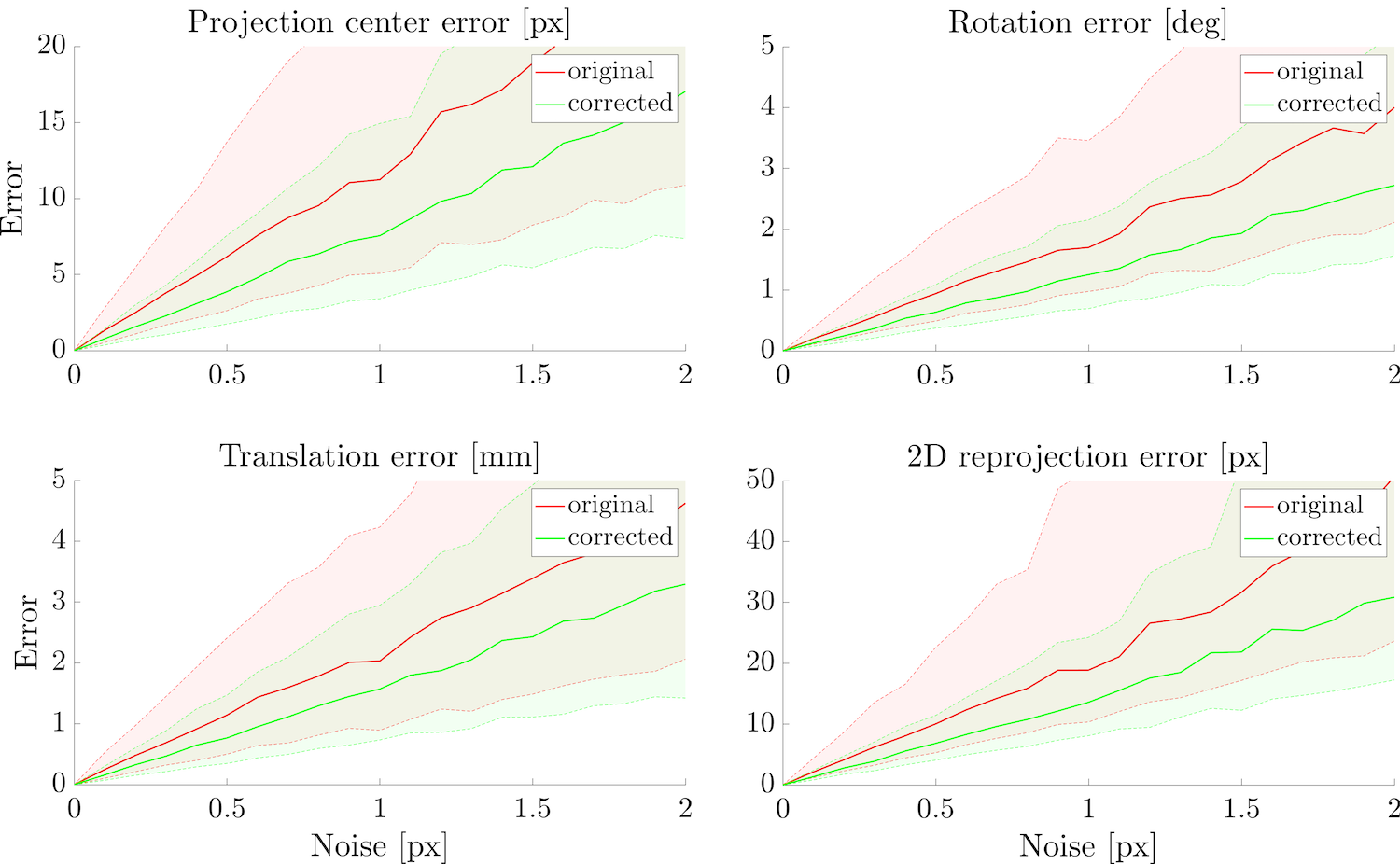}
\caption{\textbf{Correcting corners improves the initial guess.}
  We evaluate the accuracy of the center of projection, camera pose
  and rewarped points using the original and corrected
  corners. Evaluation is done over 1000 experiments at each noise
  level. Solid curves are median errors, and shaded regions are
  interquartile ranges.}
\label{fig:correction}
\end{center}
\vspace{-25pt}
\end{figure}

\input{corner_correction}

\subsection{Solving the Remaining Intrinsics and Depth}
The homography $\mH\xspace$ mapping from the camera coordinate system
to coordinates of the retinal plane
can be used to solve for the remaining
parameters, 
$
\gamma g(\mA\vu) = \diag{f,f,1}\mH\vx.
$
Note that $s$ and $f$ in $\mA$
cannot be disentangled without additional knowledge about the camera such as the pixel size. Further, we assume that this information is unavailable, and let $f \leftarrow sf$. An unknown $\gamma$
is eliminated through the cross product,
\begin{equation}
  \label{eq:intrinsics_constraints1}
  g\left(\diag{\inv{f},\inv{f},1}\vu^{\prime}\right) \times \colvec{3}{x^{\prime}}{y^{\prime}}{z^{\prime}+t_z} = 0,
\end{equation}
where $\vu^{\prime} = \diag{\inv{a},1,1}\ma{T}(-\ve) \vu$, $x^{\prime} = \vh_1 \tr \vx$, $y^{\prime} = \vh_2\tr \vx$, and $z^{\prime} = \rowvec{3}{r_{31}}{r_{32}}{0} \vx$. Reparameterizing $\tilde\lambda_k = \lambda_k/f^{2k-1}$ and collecting terms in \eqref{eq:intrinsics_constraints1} gives a system linear in the unknowns 
\begin{equation}
    \begin{bmatrix}
        & & \vdots \\
      x_i^{\prime} & x_i^{\prime} {r_i^{\prime}}^2 & ... & x_i^{\prime} {r_i^{\prime}}^{2N} & -u_i^{\prime} \\
      y_i^{\prime} & y_i^{\prime} {r_i^{\prime}}^2 & ... & y_i^{\prime} {r_i^{\prime}}^{2N} & -v_i^{\prime} \\
        & & \vdots \\
    \end{bmatrix}
    \begin{pmatrix}
        f \\ \tilde{\lambda}_1 \\ \vdots \\ \tilde{\lambda}_N\\ t_z
    \end{pmatrix} = 
    \begin{pmatrix}
        \vdots \\
        u_i^{\prime} \cdot z_i^{\prime} \\
        v_i^{\prime} \cdot z_i^{\prime} \\
        \vdots \\
    \end{pmatrix}
    \label{eq:intrinsic_solution}
\end{equation}
where $r_i^{\prime} = r(\vu_i^{\prime})$ is the radius of the point $\vu_i^{\prime}$.
\vspace{-5pt}

\paragraph{Model Selection for the Division Model}
The degree of division model $\psi(\cdot)$ defined in
\eqref{eq:bproj_division_model} needs to be chosen such that it can
approximate the extreme radial profiles of fisheye and catadioptric
rigs, and so that it does not overfit to noisy measurements for narrow
field-of-view lenses. Clearly these are competing goals. We evaluated
$\psi(\cdot)$ for polynomials of even degrees from two to ten. Model
selection is performed on the dataset introduced in
\secref{sec:evaluation}, which contains a wide range of lenses as well
as catadioptric rigs.

The camera model of \eqref{eq:omni_camera_model} is estimated linearly
from sampled corner correspondences (denoted \emph{Initial} in
\tabref{tab:solver-complexity}) as outlined above and fit with
non-linear least squares (denoted \emph{Refined} in
\tabref{tab:solver-complexity}) using all corners for each calibration
capture. The weighted RMS reprojection error and inlier ratio are used
to assess the accuracy of each model's initial guess and refined
solution across the entire dataset.
\tabref{tab:solver-complexity} shows that models of degrees eight and
ten significantly deviate from the optimal result, suggesting that
they are over-fitting. The fourth-degree division model parameterized
by $\{\,\lambda_1, \lambda_2\,\}$ is the simplest model that is
sufficiently flexible to provide a good initial guess. We estimate the
back-projection function \eqref{eq:omni_camera_model} using
\eqref{eq:bproj_division_model} with $N=2$.

\begin{table}[!t]
\begin{center}
\footnotesize
\begin{tabular*}{0.478\textwidth}{lrr}
\toprule
Degree & Initial RMS [px],\,\, inl. [\%] & Refined RMS [px],\,\, inl. [\%] \\
\midrule
2  & 5.342,\,\,\, 26.042 & 0.647,\,\,\, 97.086 \\
\textbf{4}  & \sb{4.585},\,\,\, \fb{28.991} & \fb{0.587},\,\,\, \fb{97.403} \\
6  & \fb{4.516},\,\,\, \sb{26.184} & \sb{0.587},\,\,\, \sb{97.399} \\
8  & 7.429,\,\,\, 13.307    & 1.020,\,\,\, 93.660  \\
10 & 12.804,\,\,\, \,\,\,8.448    & 4.812,\,\,\, 61.267  \\
\bottomrule
\end{tabular*}
\caption{\textbf{Model selection for the division model.} A polynomial of degree four gives the best results overall.
} 
\label{tab:solver-complexity}
\end{center}
\vspace{-25pt}
\end{table}

\vspace{-5pt}
\paragraph{Model-to-Model Regression}
 A radial projection function, denoted $\phi_\theta(R,Z)$, can always
 be parameterized by how it maps a point at radius $R$ from the
 optical axis and depth $Z$ from the principal plane to the retinal
 plane (see \tabref{tab:models}). This parameterization admits a
 universal way to regress radially-symmetric projection functions
 against the division model.

If the user-selected projection model does not have the division model
for its radial profile, then the following optimization must be performed
\begin{equation}
    \sum_k(\phi_\theta(r_k,\psi(r_k)) - r_k)^2 \rightarrow
    \min_\theta,
\label{eq:model_to_model_regression}
\vspace{-7pt}
\end{equation}
where radii  $r_k = \frac{k-1}{K-1} r^{\max}$ are uniformly sampled
from zero to the maximum radius $r^{\max}$.
All regressions for the forward models in \tabref{tab:models} are either linear (BC, KB, UCM, EUCM) or easy to perform with the least squares (FOV, DS). See \secref{sec:supp_m2m_regression} in Supplemental for the details and example of regression against the Kannala-Brandt model.

%% file: corner_correction.tex
\subsection{Corner Correction}
Corner correction is defined such that given the radial fundamental
matrix $\mF_r$ and correspondence $\cspond{\vu_i}{\vx[i]}$, the
corrected corner is $\vu_i^*=\vu_i+\delta_{\vu_i}$, where $\delta_{\vu_i}$
is the smallest displacement such that $\vu_i^*$ satisfies the epipolar
constraint $\vu_i^{*\T}\mF_r\vx[i]=0$. The target fiducials $\{\,\vx[i]
\,\}$ are assumed correct since they are noiseless. It can be shown
\cite{Hartley-BOOK04} that the corrected corner $\opt{\vu_i}$ is
recovered by projecting the measured corner $\vu_i$ onto the epiline
of $\vx[i]$,
\begin{equation}
 \label{eq:corner_correction} \opt{\vu_i}
 = \mbox{proj}_{\mF_r\vx[i]}(\vu_i).
\end{equation}

We refine the radial fundamental matrix $\mF_r$ by minimizing the
displacements with non-linear least squares. Eight correspondences are
sufficient for correcting the sampled corners \cite{Hartley-BOOK04},
but it is reasonable to use more since we expect a high inlier ratio
for a calibration capture. The rank-two constraint is encoded with the
parameterization $\mF_r = \sksym{\ve}
\rowvec{3}{\vh_1^{\T}}{\vh_2^{\T}}{\vzero_3}\tr$. Then the refined
fundamental matrix is recovered by solving
\begin{equation}
\label{eq:geometric_correction_objective}
\ve^*,\vh^*_1,\vh^*_2 =  \argmin_{\ve,\vh_1,\vh_2} \sum_i \delta_{\vu_i}\tr \delta_{\vu_i} 
\end{equation}
and reconstructing $\mF^*_r$ from $\ve^*,\vh^*_1,\vh^*_2$. The noisy
detected corners are corrected according
to \eqref{eq:corner_correction} using $\mF^*_r$.

\paragraph{Evaluation of Corrected Corners}
Synthetic scenes were used to measure the accuracy gains to camera
model estimation from corner correction. The camera was randomly posed
to view a chessboard. Image resolution was $1200 \times 800$ pixels,
focal length $400$ pixels, and the center of projection was displaced
to $\rowvec{2}{700}{500}$.  We added different levels white noise to
the corners: $\sigma \in \buildset{0,0.1,0.2,...,2}{}{}$. Camera
models were fit using either the original or corrected corners for
1000 images.

Clockwise from the top left, \figref{fig:correction}
reports \begin{enumerate*}[(i)] \item the distance between the
  estimated and ground truth center-of-projection, \item the smallest
  angle of rotation required to correct the estimated orientation,
  \item the distance between the estimated and ground-truth camera
    position, and \item the RMS reprojection error between an image
    grid and the reprojection of scene points by the estimated camera
    that should project onto the image
    grid \end{enumerate*}. \figref{fig:correction} shows that
correcting corners reduces median errors of rotation, translation and
RMS reprojection error by $31\%$, $28\%$, and $33\%$ on average.

%% file: method.tex
\begin{figure*}[!t]
\begin{center}
\scriptsize
\begin{tabular}{c c c}
\ovplane---\texttt{130108MP},\,\, $0.478$ px RMS & \ovcorner---\texttt{Cam4},\,\, $0.770$ px RMS & \ovcube---\texttt{Cam1},\,\, $0.268$ px RMS\\
\includegraphics[width=0.31\textwidth]{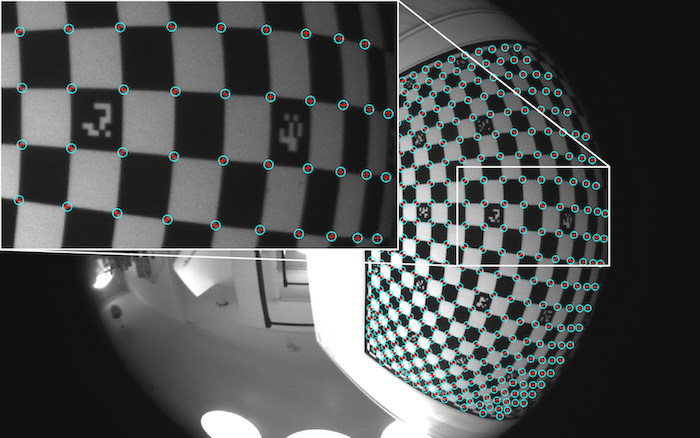} &
\includegraphics[width=0.31\textwidth]{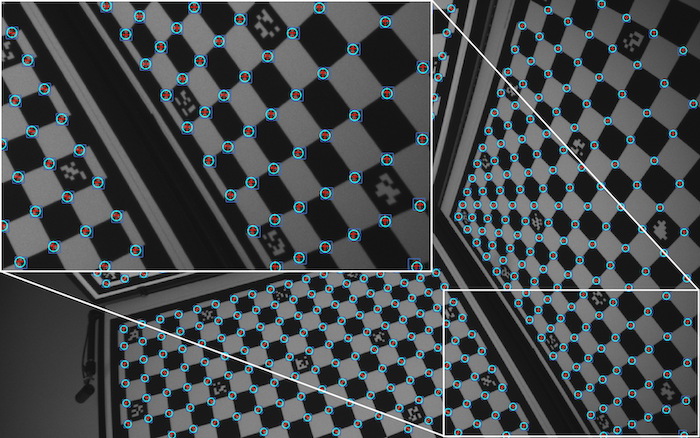} &
\includegraphics[width=0.31\textwidth]{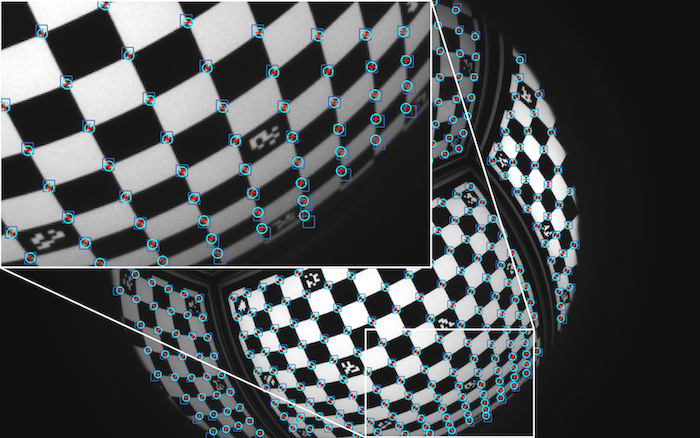} \\
\end{tabular}
\vspace{-10pt}
\caption{\textbf{Projection of calibration target from estimated calibration.}
Detected corners are red crosses, target projected using initial
calibration are blue squares and using the final calibration are cyan
circles.}
\label{fig:OV-results}
\end{center}
\vspace{-20pt}
\end{figure*}

\section{Robust Estimation Framework}
\label{sec:robust_estimation}

In this section we propose a calibration framework that is robust to
corner detection errors, works with either one or multiple calibration
boards, and handles the partial visibility of board fiducials across
the calibration capture. The robustness of the method is, in part,
achieved using the camera geometry estimators proposed
in \secref{sec:good_guess}. With these estimators, an accurate
estimate of any of the board models listed in \tabref{tab:models} can
be recovered from a sample of noisy corner detections. However, corner
extraction can fail due to the detector's inability to localize
highly-distorted saddle points \cite{Ha-ICCV17}. The grid search can
also fail at the extents of a fisheye image because of
highly-distorted neighborhoods.  Occlusions can also create false
corners.
We incorporate the solvers of \secref{sec:good_guess} into
a \RANSAC-based framework to handle bad
detections \cite{Fischler-CACM81}. The method fits camera models
sampled from corner-to-board correspondences. The models are scored
with a robust objective. During sampling, the best-so-far
model is kept and refined with a maximum-likelihood estimator, which
is inspired by the local optimization step in \cite{Chum-PR03}. The
output of the model is a maximum-likelihood fit of camera intrinsics
and extrinsics. The estimators proposed in \secref{sec:good_guess} are
very simple, so they are fast and are well-suited for use in the model
proposal step of \RANSAC.
\algref{alg:method} in \secref{sec:supp_algorithm} of Supplemental may be helpful as a
cross-reference for the next paragraphs that specify the method.

The input to the method is a set of 2D-3D correspondences that match
corner detections with board fiducials, which we denote
$\cspond{ \vX[ij] }{ \vu_{ijk} }$. Indices $i$ and $j$ indicate a
particular fiducial $i$ on plane $j$, and $k$ indicates the image of
the corner detection $\vu_{ijk}$ of fiducial $\vX[ij]$. For
each \RANSAC iteration, we sample an image $k$, a plane $j$ visible in
image $k$, and a non-minimal sample of $14$ correspondences that are
used to compute the radial fundamental matrix, center of projection,
and corner correction according to \eqref{eq:billinear_form},
\eqref{eq:principal_point}, and \eqref{eq:corner_correction}.
The utilized sample size of $14$ correspondences was cross-validated.

\begin{table}[!t]
\begin{center}
\footnotesize
\begin{tabular*}{0.478\textwidth}{llrr}
\toprule
Dataset & \# cam., \# img. & DFOV range & \hspace*{-2pt}Max. img. size\\
\midrule
\kalibrdata & $8$,\, $140+60$ & $110\deg$---$268\deg$ & $1680\times 1680$ \\
\occdata & $9$,\, $\,\,\, 79+40$ & $130\deg$---$266\deg$  & $3840\times2880$ \\
\uzhdavis & $4$,\, $140+60$ & $124\deg$---$148\deg$ & $346\times260$ \\
\uzhsnap\hspace*{-9pt}& $4$,\, $140+60$ & $144\deg$---$166\deg$ & $640\times480$ \\
\hline
\ovcorner & $8$,\, $280+120$ & $109\deg$---$109\deg$ & $1280\times800$ \\
\ovcube & $4$,\, $105+49$ & $159\deg$---$183\deg$ & $1280\times800$ \\
\ovplane & $4$,\, $\,\,\, 92+41$ & $88\deg$---$187\deg$ & $1280\times800$\\
\bottomrule
\end{tabular*}
\normalsize
\caption{\textbf{Calibration dataset details.}  Train-test split of the
  images is indicated by $+$. The diagonal field of view (DFOV) is
  approximated using intrinsic calibration.}
\label{tab:datasets}
\end{center}
\vspace{-25pt}
\end{table}

The aspect ratio is a necessary parameter for the pose and intrinsics
estimators. If the camera has non-square pixels or an anamorphic lens, we sample an aspect ratio from the interval $[0.5,2]$. Pose and
intrinsics are estimated as derived
in \eqref{eq:orthonormality_constraints}
and \eqref{eq:intrinsic_solution}. The radial profile of the user-selected camera model is regressed against the radial profile of
the division model using \eqref{eq:model_to_model_regression}, if the
division model is not the desired model. The model-to-model regression
generates the camera geometry portion of the \RANSAC model
proposal. Given the estimate of intrinsics, the poses for the remaining
images are computed using P3P (Perspective-3-Point \cite{Nakano-BMVC-2019})
from three sampled corner-to-board correspondences. The camera poses are added 
to the intrinsics to give a \RANSAC model proposal.

\begin{table}[!t]
\begin{center}
\footnotesize
\begin{tabular*}{0.478\textwidth}{lrrr}
\toprule
& \opencvBC & \kalibrBC &\ourBC \\
\midrule
& \multicolumn{3}{c}{RMS [px],\,\, inl. [\%]}  \\
\midrule
\texttt{Cam0}\hspace*{10pt} &\hspace*{10pt} 0.886,\,\, 90.9 &\hspace*{10pt} 0.945,\,\, 87.3&\hspace*{10pt} \ffb{0.704,\,\, 96.1}  \\
\texttt{Cam1} & 0.781,\,\, 95.2 & 0.893,\,\, 88.8& \ffb{0.674,\,\, 98.0}  \\
\texttt{Cam2} & 0.773,\,\, 96.1 & 0.756,\,\, 95.6& \ffb{0.720,\,\, 96.9}  \\
\texttt{Cam3} & 0.733,\,\, 97.0 & 0.953,\,\, 87.6&      0.710,\,\, 96.4  \\
\texttt{Cam4} & 0.757,\,\, 97.3 & 0.828,\,\, 93.6& \ffb{0.679,\,\, 97.6}  \\
\texttt{Cam5} & 0.772,\,\, 96.4 & 0.831,\,\, 91.7&      0.759,\,\, 96.0  \\
\texttt{Cam6} & 0.715,\,
\, 95.4 & 0.748,\,\, 94.5& \ffb{0.677,\,\, 96.0}  \\
\texttt{Cam7} & 0.701,\,\, 96.6 & 0.855,\,\, 90.6& \ffb{0.641,\,\, 97.5}  \\
\bottomrule
\end{tabular*}
\caption{\textbf{Pose evaluation for the BC model.} \ovcorner test images used.}
\label{tab:holdout-test-bc}
\end{center}
\vspace{-25pt}
\end{table}

\begin{table*}
\begin{center}
\footnotesize
\begin{tabular*}{\textwidth}{lrrrrrr}
\toprule
               & \opencvKB          & \kalibrKB               & \ourKB               & \opencvUCM       & \kalibrUCM      & \ourUCM\\
\midrule
& \multicolumn{6}{c}{RMS [px],\,\, inl. [\%],\,\, \# failures}  \\
\midrule
\ovcorner   & 0.746,\,\, 94.4,\,\, 0/8 & 0.882, 90.3,\,\, 0/8 & \ffb{0.695,\,\, 96.8,\,\, 0/8} & 1.187,\,\, 75.4,\,\, 0/8  & 0.953,\,\, 88.8,\,\, 0/8 & \ffb{0.812,\,\, 94.8,\,\, 0/8} \\
\ovcube     &           FAIL,\,\,  4/4 & 0.411, 92.7,\,\, 2/4 & \ffb{0.265,\,\, 97.5,\,\, 0/4} & 0.493,\,\, 86.7\,\,  0/4  & 0.440,\,\, 90.8\,\,  0/4 & \ffb{0.316,\,\, 96.8,\,\, 0/4} \\
\ovplane    & 2.449,\,\, 70.2,\,\, 0/4 & 0.658, 92.9,\,\, 1/4 & \ffb{0.596,\,\, 94.0,\,\, 0/4} & 0.854,\,\, 80.7,\,\, 0/4  & 0.669,\,\, 90.6,\,\, 0/4 & \ffb{0.606,\,\, 93.6,\,\, 0/4} \\
\kalibrdata & 2.291,\,\, 53.9,\,\, 3/8 & 0.194, 99.8,\,\, 3/8 & \ffb{0.173,\,\, 99.9,\,\, 0/8} & 0.355,\,\, 95.2,\,\, 0/8  & 0.350,\,\, 94.6,\,\, 0/8 & \ffb{0.326,\,\, 97.1,\,\, 0/8} \\
\occdata    & 2.921,\,\, 67.0,\,\, 3/9 & 0.696, 97.2,\,\, 4/9 & \ffb{0.676,\,\, 97.3,\,\, 0/9} & 0.782,\,\, 93.4,\,\, \red{2/9}  & 0.776,\,\, 94.6,\,\, 0/9 &      0.784,\,\, 97.2,\,\, \green{0/9}  \\
\uzhdavis   & 0.389,\,\, 96.3,\,\, 0/4 & 0.389, 96.3,\,\, 0/4 & \ffb{0.382,\,\, 96.3,\,\, 0/4} & 0.503,\,\, 95.3,\,\, 0/4  & 0.490,\,\, 93.2,\,\, 0/4 & \ffb{0.385,\,\, 96.2,\,\, 0/4} \\
\uzhsnap\hspace*{10pt}    & 0.265,\,\, 99.6,\,\, 0/4 & 0.268, 99.6,\,\, 0/4 & \ffb{0.254,\,\, 99.6,\,\, 0/4} & 0.517,\,\, 97.3,\,\, 0/4  & 0.299,\,\, 99.4,\,\, 0/4 & \ffb{0.286,\,\, 99.5,\,\, 0/4} \\
\bottomrule
\toprule
            & \kalibrFOV & \ourFOV & \kalibrEUCM  & \ourEUCM & \kalibrDS  & \ourDS\\
\midrule
& \multicolumn{6}{c}{RMS [px],\,\, inl. [\%],\,\, \# failures}  \\
\midrule
\ovcorner   & 0.931,\,\, 88.7,\,\, 0/8 & \ffb{0.743,\,\, 96.1,\,\, 0/8} &            FAIL,\,\, 8/8 & \ffb{0.751,\,\, 95.8,\,\, 0/8} &      0.967,\,\, 88.5,\,\, 0/8  & \ffb{0.812,\,\, 94.5,\,\, 0/8} \\
\ovcube     &            FAIL \,\, 4/4 & \ffb{1.356,\,\, 19.3,\,\, 0/4} & 0.416,\,\, 91.9,\,\, 3/4 & \ffb{0.273,\,\, 97.0,\,\, 0/4} &      0.413,\,\, 92.7,\,\,  0/4  & \ffb{0.269,\,\, 97.3,\,\, 0/4} \\
\ovplane    & 0.867,\,\, 83.1,\,\, 1/4 & \ffb{0.863,\,\, 82.8,\,\, 0/4} & 0.584,\,\, 96.8,\,\, \red{2/4} &      0.542,\,\, 96.6,\,\, \green{0/4}  &      0.644,\,\, 93.2,\,\, 0/4  & \ffb{0.606,\,\, 93.6,\,\, 0/4} \\
\kalibrdata & 0.257,\,\, 99.1,\,\, 3/8 & \ffb{0.237,\,\, 99.2,\,\, 0/8} & 0.250,\,\, 98.9,\,\, 0/8 & \ffb{0.230,\,\, 99.2,\,\, 0/8} & \ffb{0.264,\,\, 98.2,\,\, 0/8} &      0.326,\,\, 97.1,\,\, 0/8  \\
\occdata    & 0.786,\,\, 96.1,\,\, 4/9 & \ffb{0.779,\,\, 96.2,\,\, 0/9} & 0.580,\,\, 97.8,\,\, \red{6/9} &      0.561,\,\, 97.7,\,\, \green{0/9}  &      0.755,\,\, 94.6,\,\, 0/9  & \ffb{0.739,\,\, 97.7,\,\, 0/9} \\
\uzhdavis   & 0.421,\,\, 95.5,\,\, 1/4 & \ffb{0.417,\,\, 95.7,\,\, 0/4} & 0.415,\,\, 95.6,\,\, 1/4 & \ffb{0.411,\,\, 95.6,\,\, 0/4} &      0.393,\,\, 96.2,\,\, 0/4  & \ffb{0.382,\,\, 96.2,\,\, 0/4} \\
\uzhsnap    & 0.250,\,\, 99.6,\,\, 1/4 & \ffb{0.234,\,\, 99.6,\,\, 0/4} & 0.246,\,\, 99.6,\,\, 1/4 & \ffb{0.232,\,\, 99.6,\,\, 0/4} &      0.284,\,\, 99.3,\,\, 0/4  &      0.286,\,\, 99.5,\,\, 0/4  \\
\bottomrule
\end{tabular*}
\caption{\textbf{Pose Evaluation for fisheye and catadioptric rigs.}
  Estimated models are (top) KB, UCM, and (bottom) FOV, EUCM, and DS.
  Each method's performance on a dataset is given by the weighted
  RMS reprojection error [px], the inlier ratio [\%], and the number
  of catastrophic failures.
}
\label{tab:holdout-test}
\end{center}
\vspace{-25pt}
\end{table*}

The reprojection error is evaluated against the entire calibration
capture with the robust objective
\begin{equation}
    \mathcal{J}(\Theta) = \sum_{ijk} \rho(d(\pi\left(\begin{bmatrix}\mR_{jk}
    & \vt_{jk} \end{bmatrix} \vX[ij] ), ~ \vu_{ijk})\right),
\label{eq:robust_loss_pose}
\end{equation}
where $\pi(\cdot)$ is the selected projection function,
$d(\cdot,\cdot)$ is the Euclidean distance, $\rho(\cdot)$ is the the
Huber loss function \cite{Huber-BOOK04}, and $\Theta =
\{\theta,\mK,\mR_{jk},\vt_{jk}\}$ are the calibration parameters. In
the case of multiple planar targets, the poses $\mR_{jk}, \vt_{jk}$ are
constructed using the absolute poses of the cameras, $\mR^c_{k},
\vt^c_{k}$, and the relative poses of the boards with respect to the
reference board, $\mR^b_{j}, \vt^b_{j}$,
\begin{equation*}
\mR_{jk} = \mR^c_{k} \mR^b_{j} \qquad
\vt_{jk} = \mR^c_{k} \vt^b_{j} + \vt^c_{k}.
\end{equation*}

If \RANSAC encounters a best-so-far calibration proposal, then the
model refinement step, $\mathcal{J}(\Theta)\to\min_{\Theta}$, is
invoked. The axis-angle representation is used to minimally parameterize the
rotations for the bundle adjustment. Proposals are ranked by their inlier ratio, and the inlier
ratio is computed according to
\begin{equation}
\mathcal{I}(\Theta) = \frac1M
\sum_{ijk} 1_{\{\leq \tau\}}\left(d\left(\pi\left(\begin{bmatrix}\mR_{jk}
~ \vt_{jk} \end{bmatrix} \vX[ij]\right), \vu_{ijk}\right)\right),
\label{eq:inlier_ratio}
\end{equation}
where $M$ is the total number of image-to-target correspondences,
$1_A\left(\cdot\right)$ is an indicator function,
and $\tau$ is the scale of the robust estimator.

%% file: experiments.tex
\section{Evaluation}
\label{sec:evaluation}
The benchmark surveys a wide variety of lens types and includes
catadioptric rigs. We aggregate several established datasets that are
commonly used for testing the accuracy of camera calibration
frameworks: \begin{enumerate*}[(i)] \item Double Sphere
  \cite{Usenko-3dv-2020}, \item EuRoC \cite{Burri-IJRR-2016}, \item
  TUM VI \cite{Schubert-IROS-2018}, \item and ENTANIYA~\footnote{found
  on github:
  https://github.com/ethz-asl/kalibr/issues/242}\end{enumerate*}.  We
call the aggregated dataset \kalibrdata since the Kalibr calibration
framework \cite{Maye-IVS-2013} was used in the original publications
cited above. \kalibrdata has eight cameras, most of which are
fisheyes.  We also test the \occdata \cite{Scaramuzza-ICVS-2006}
dataset, which has five fisheye lenses and four catadioptric rigs; and
the \texttt{UZH} \cite{Delmerico-ICRA-2019} dataset, which consists of
eight wide-angle and fisheye cameras captured from a drone. The
dataset is split into two subsets\footnote{according to:
https://fpv.ifi.uzh.ch/datasets/}, \texttt{DAVIS} and
\texttt{Snapdragon}.

We also acquired calibration data from sixteen OmniVision cameras
fitted with different lenses giving fields of view ranging from
$88\deg$ to $187\deg$. The cameras were calibrated using three
different targets containing AprilTags \cite{Olson-ICRA-2011}:
\texttt{Plane}, \texttt{Corner} and \texttt{Cube}, with one, three,
and four chessboards, respectively (see \figref{fig:OV-results}). The
OmniVision capture is denoted \texttt{OV}. The specifications of the
cameras from each dataset are listed in \tabref{tab:datasets}.

\begin{table}[!b]
\begin{center}
\footnotesize
\begin{tabular*}{0.478\textwidth}{l@{\extracolsep{\fill}}rr}
\toprule
          & \occDiv  &\ourDiv \\
\midrule
& \multicolumn{2}{c}{RMS [px],\,\, inl. [\%]}  \\
\midrule
\texttt{Fisheye1}           &        0.631,\,\, 98.3 &      0.603,\,\, 97.1 \\
\texttt{Fisheye190deg}      &        0.642,\,\, 96.8 &      0.621,\,\, 95.2 \\
\texttt{Fisheye2}           &        0.480,\,\, 97.9 & \ffb{0.458,\,\, 97.9}\\
\texttt{GOPRO}              &        1.097,\,\, 95.3 &      1.177,\,\, 96.9 \\
\texttt{KaidanOmni}         &      0.595,\,\, 100\,  &      0.574,\,\, 98.3 \\
\texttt{Ladybug}            &        0.661,\,\, 98.8 &      0.658,\,\, 97.5 \\
\texttt{MiniOmni}           &        0.795,\,\, 97.7 &      0.712,\,\, 95.6 \\
\texttt{Omni}               &        0.828,\,\, 93.3 &      0.836,\,\, 97.1 \\
\texttt{VMRImage}           & \ffb{0.560,\,\, 100\,} &      0.560,\,\, 99.2 \\
\bottomrule
\end{tabular*}
\caption{\textbf{Pose evaluation for the DIV model}. The test images from \occdata were used.}
\label{tab:holdout-test-sc}
\end{center}
\vspace{-25pt}
\end{table}

We evaluated state-of-the-art camera calibration frameworks that
support the projection models listed in \tabref{tab:models}. OpenCV
supports three models: BC~\cite{Brown-P-1971},
KB~\cite{Brandt-PAMI-2006}, and
UCM~\cite{Mei-ICRA-2007,Li-ICIRS-2013}. In addition to those models,
the Kalibr framework~\cite{Maye-IVS-2013} supports the FOV
\cite{Devernay-MVA01}, EUCM \cite{Khomutenko-RAL15} and
DS~\cite{Usenko-3dv-2020} models.  The OCamCalib (MATLAB) framework
for fisheye and catadioptric rigs has the
DIV~\cite{Scaramuzza-ICVS-2006,Urban-ISPRS-2015} model.

BabelCalib can regress all the camera models listed in
\tabref{tab:models}.  We compare each state-of-the-art calibration
method over their supported models with the BabelCalib estimation of
the same models. The state of the art was provided with a reasonable
initial guess for the focal length, and the center of projection was
initialized to the image center. The proposed BabelCalib does not
require, nor was it given, a user-provided initial guess.

\begin{figure*}[!t]
\begin{center}
\footnotesize
\begin{tabular*}{\textwidth}{c c c c}
\textsc{\qquad Original} & \textsc{\quad Displaced} & \textsc{\quad Non-Square} & \textsc{\quad Displaced + Non-Square} \\
\includegraphics[height=0.2\textwidth]{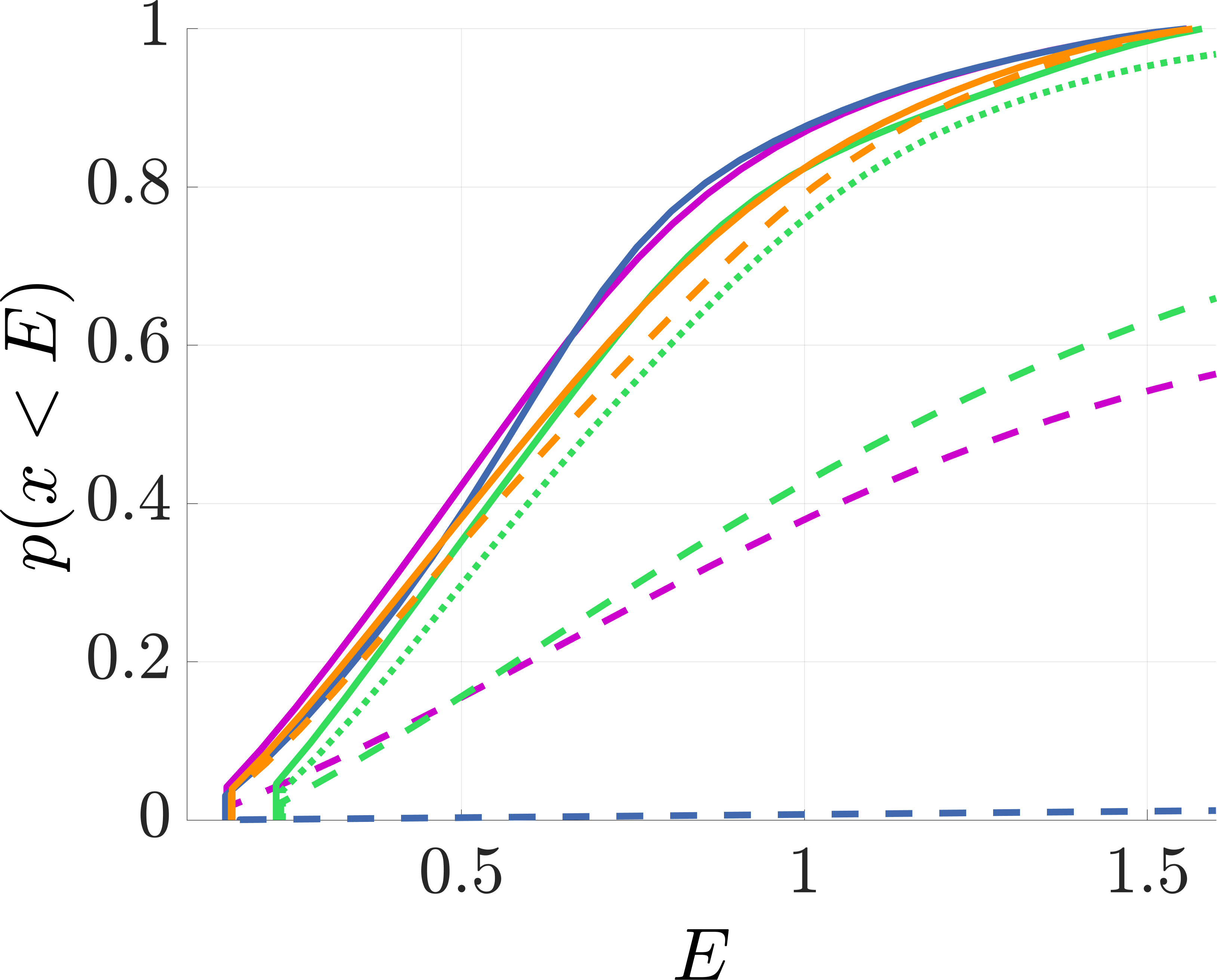} &
\includegraphics[height=0.2\textwidth]{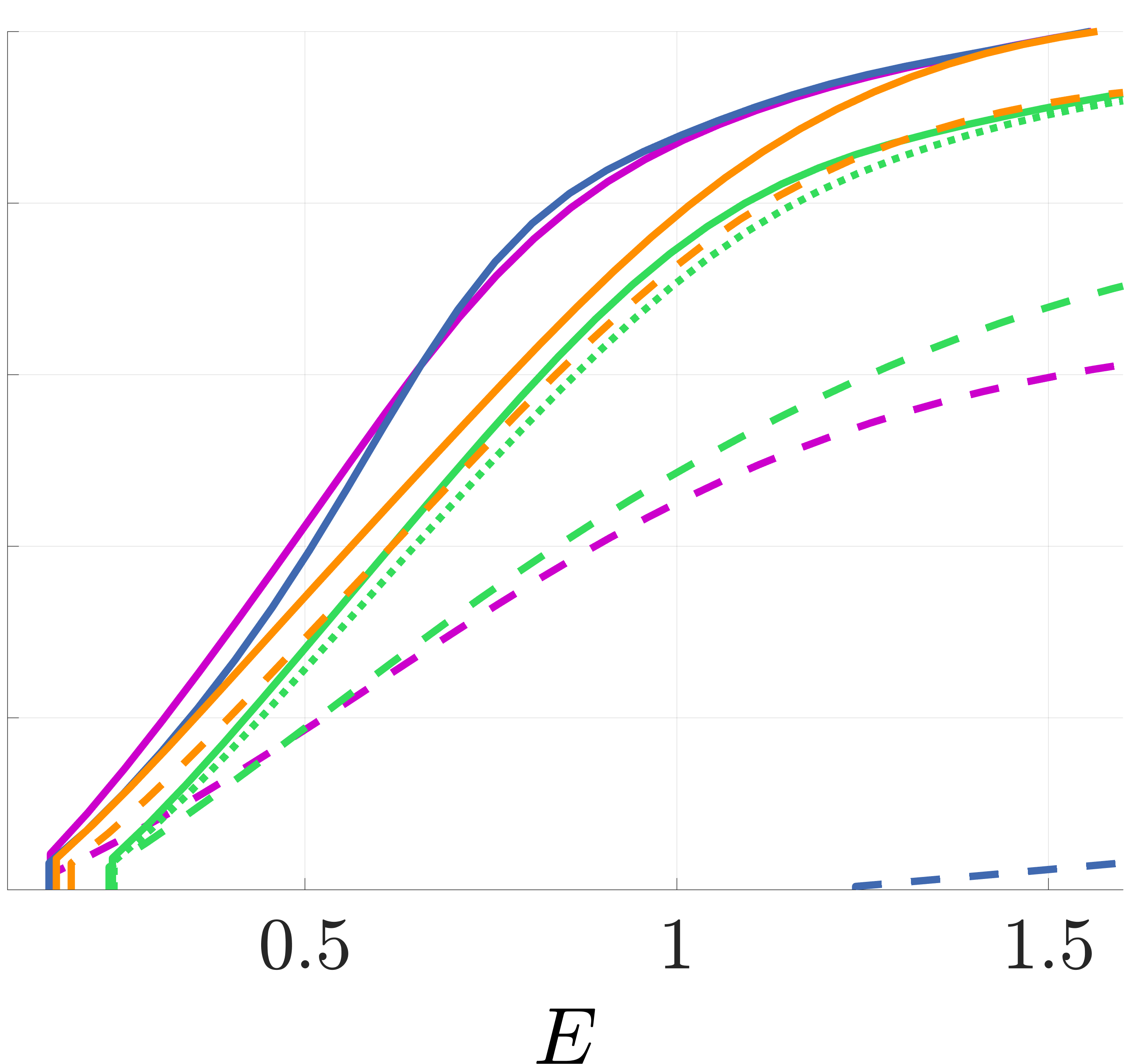} &
\includegraphics[height=0.2\textwidth]{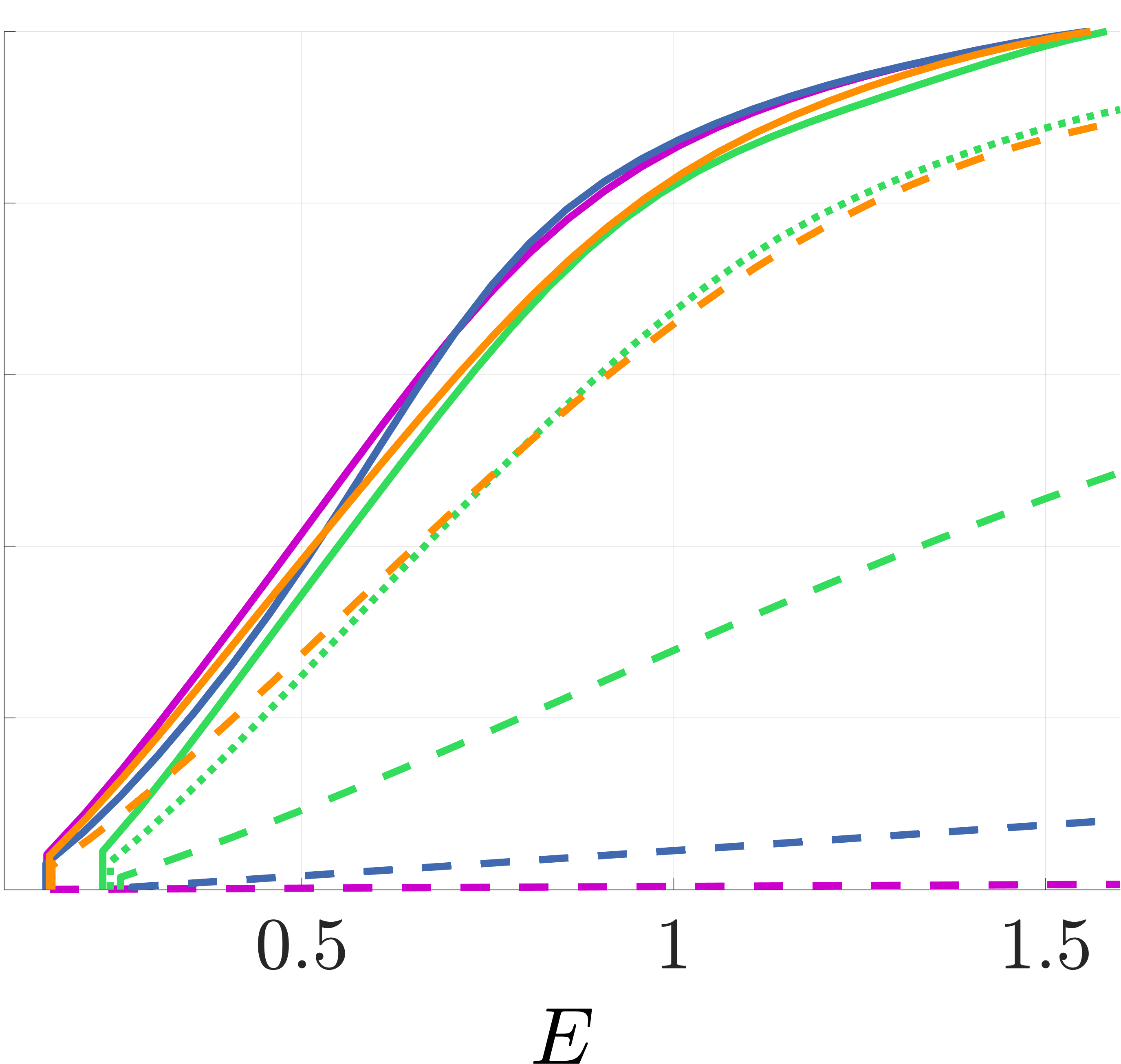} &
\includegraphics[height=0.2\textwidth]{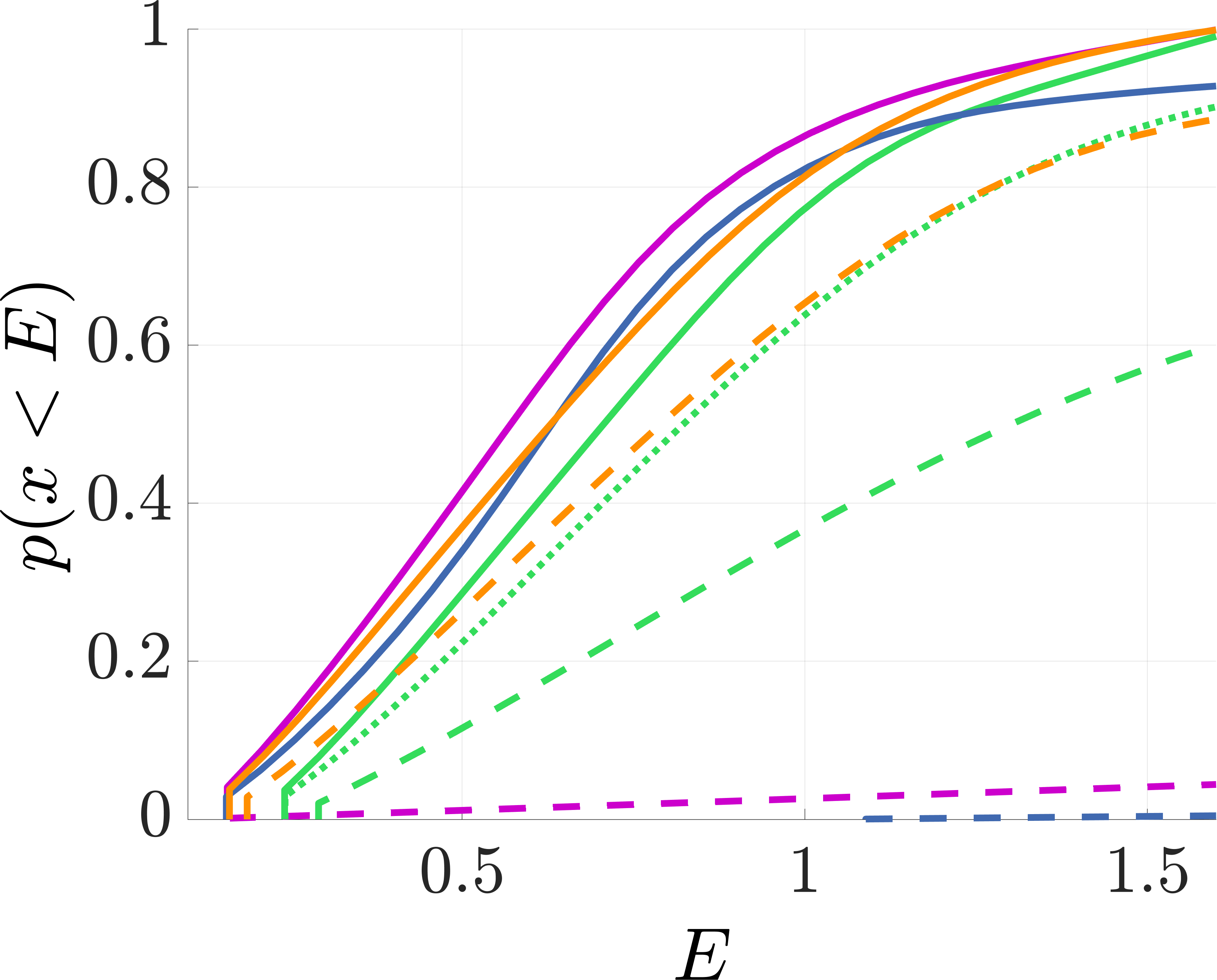} \\
\end{tabular*}
\includegraphics[height=0.05\textwidth]{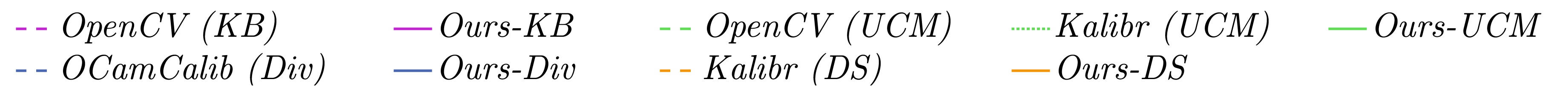}
\caption{\textbf{Performance summary for all calibration captures}. The
  empirical CDFs show the relative performance of each method across
  all calibration datasets. The methods are tested for different
  augmentation types.  The probability that the weighted RMS
  reprojection error is less than $E$ is plotted. The weighted RMS
  reprojection error is normalized to correspond to a $1000\times1000
  $ px image. BabelCalib performs better for every model type on the
  original and augmented imagery.}
\label{fig:holdout-cdfs}
\end{center}
\vspace{-25pt}
\end{figure*}

\subsection{Camera Pose Estimation}
Pose accuracy for held-out test images from the calibration captures
is used to evaluate the calibration of each method. Camera intrinsics
are fixed to the calibration, and the objective $\mathcal{J}(\Theta)$
of \eqref{eq:robust_loss_pose} is minimized over camera poses only. The
train-test split for each dataset is specified in
\tabref{tab:datasets}. Pose accuracy is measured by the robust RMS
reprojection error of \eqref{eq:robust_loss_pose} and the mean inlier
ratio evaluated by \eqref{eq:inlier_ratio}. The error and inlier ratio
for each dataset are averaged across the cameras where the method
returns a model. This policy favors the state of the art since
BabelCalib returns models for all cameras. These measures are reported
for each framework-dataset combination in the comparison tables.  The
winner is boldfaced. A tie is declared if a method is not best on both
measures. In this, we mark the highest number of failures in red.

\begin{table}[!b]
\begin{center}
\footnotesize
\begin{tabular*}{0.45\textwidth}{lr|lr}
\toprule
{\opencvBC} & 8.76\% &
{\kalibrBC} & 17.68\%\\
{\opencvKB} & 42.96\%&
{\kalibrKB} & 12.41\%\\
{\opencvUCM} & 24.66\% &
{\kalibrUCM} & 12.00\% \\
{\kalibrFOV} & 6.11\% &
{\kalibrEUCM} & 9.92\% \\
{\occDiv} & 2.05\% &
{\kalibrDS} & 5.36\% \\
\bottomrule
\end{tabular*}
\caption{\textbf{Accuracy gains with respect to model type.}
  BabelCalib regressions significantly reduce the weighted RMS
  reprojection error for all camera model types with respect to the
  estimates provided by the tested frameworks.}
\label{tab:reduction}
\end{center}
\end{table}

\tabref{tab:holdout-test-bc} reports results for the narrow-medium FOV
lenses in the \ovcorner dataset.  We breakout the results for each
camera. The suitable model for this lens type is a pinhole projection
with additive BC distortion. BabelCalib outperforms OpenCV and
Kalibr. \tabref{tab:holdout-test} reports results for fisheye lenses and
catadioptric rigs.
Kalibr and OpenCV have both a 24\% failure rate for the Kannala-Brandt
model, and Kalibr has a 34\% failure rate for FOV and 51\% failure
rate for EUCM. In contrast, BabelCalib has no calibration failures.
BabelCalib consistently gives the best results for each model type,
even after discarding catastrophic failures for the state-of-the-art.

\tabref{tab:holdout-test-sc} reports calibrations using the DIV model
on the \occdata dataset. \occDiv is evaluated only on \occdata since
it requires all fiducials to be visible across the capture. \occdata
is the only dataset where this holds. BabelCalib and OCamCalib give
comparable results. This dataset includes catadioptric rigs, which
shows the flexibility of BabelCalib.

\tabref{tab:reduction} summarizes the mean reduction of RMS weighted
reprojection error for each dataset-model combination realized by
BabelCalib over the state of art. BabelCalib gives a significant error
reduction for all framework-model combinations, even after discarding
the calibration failures of the state of the art. See \figref{fig:supp_res} in Supplemental for more qualitative results and also \secref{sec:supp_limited_data} which evaluates calibration performance from a limited amount of images.

\vspace{-10pt}
\paragraph{Displaced Center and Non-Square Pixels}
We augmented the datasets by adding cropped or stretched images to
simulate a displaced projection center or a CCD with rectangular
pixels. Displacement was $\begin{pmatrix}0.15 w, 0.15 h \end{pmatrix}$
pixels for a $w \times h$ image, and pixel aspect ratio was
1.33:1. \figref{fig:holdout-cdfs} reports the distributions of robust
RMS reprojection errors for pose estimation on test images for the
original and augmented data.  Several model-framework combinations are
evaluated. For comparison, the errors are normalized to correspond to a
$1000\times1000$ pixel image. BabelCalib finds a higher percentage of
accurate calibrations on the original and augmented data for all
models.

%% file: conclusions.tex
\section{Conclusion}
BabelCalib recovers significantly more accurate calibrations than
three widely used frameworks and suffers no catastrophic
failures. BabelCalib maintains its dominance for all commonly used
models on a large survey of cameras with narrow, wide-angle and
fisheye lenses, as well as catadioptric rigs. BabelCalib maintains its
performance for cameras with displaced center of projections or
non-square pixels. It doesn't require model initialization nor
hyper-parameter tuning, so it's easy to use. Moreover, the regression
framework easily admits additional camera models.

\paragraph{Acknowledgments}
Y. Lochman, C. Zach and J. Pritts were partially supported by the Wallenberg AI, Autonomous
Systems and Software Program (WASP) funded by the Knut and Alice
Wallenberg Foundation.

%% file: supplemental.tex
\newpage
\vspace{5pt}
\begingroup
\let\cleardoublepage\relax
\let\clearpage\relax
\onecolumn
\begin{center}
\textbf{
\Large BabelCalib: A Universal Approach to Calibrating Central Cameras\\[10pt]
Supplemental Material}
\end{center}
\twocolumn
\endgroup
\normalsize
\setcounter{section}{0}
\counterwithin{figure}{section}
\counterwithin{table}{section}
\renewcommand\thesection{\Alph{section}}

\vspace{16pt}

\section{Aspect Ratio Solver}
\label{sec:supp_aspect_ratio_solver}

The epipolar constraint $\vu\tr \mF \vx = 0$ is invariant to a
projective change of coordinates. This implies that the aspect ratio
cannot be recovered using only the constraints available from the
radial fundamental matrix (\eg,
see \eqref{eq:orthonormality_constraints}).  Additional constraints
can be obtained by enforcing the concurrence of back-projected corners
with their corresponding scene points
using \eqref{eq:omni_camera_model}. The constraints given
by \eqref{eq:omni_camera_model} generate polynomial equations of the
unknown aspect ratio, remaining unknowns of $\mH$, and the unknown division model parameters
$\{a, h_{31}, h_{32}, h_{33},\lambda_1,...,\lambda_n\}$. The formulation introduces $N+1$
unknowns, but the formulation is invariant to the scale of
$\mH\xspace$. The minimal solution requires $N/2+2$ image-to-target
correspondences.

Let $\vu^{\prime}$ be the projection center-subtracted image point
$\vu^{\prime} = \ma{T}(-\ve) \vu$. Then we can
rewrite \eqref{eq:omni_camera_model} as following

\begin{equation}
\label{eq:intrinsics_constraints_aspect_ratio}
\colvec{3}{u'/f}{v'/f}{\psi(r(\diag{a^{-1},1,1}\vu'))} \times \underbrace{\diag{a,1,1}\mH}_{\hat\mH} \vx = \mathbf{0}.
\end{equation}

Solving \eqref{eq:billinear_form} and \eqref{eq:principal_point} gives the
radial fundamental matrix $\mF$ and projection center $\ve$,
respectively. We use the relation $\mF\sim\sksym{\ve}\rowvec{3}{a \vh_1\tr\vx}{\vh_2\tr\vx}{0}\tr$ to determine the
first two rows of the matrix $\hat\mH$, up to scale, $\hat\vh_1\tr
= \rowvec{3}{f_{21}}{f_{22}}{f_{23}}$ and $\hat\vh_2\tr
= -\rowvec{3}{f_{11}}{f_{12}}{f_{13}}$.

Rewriting \eqref{eq:intrinsics_constraints_aspect_ratio}
in terms of the unknowns gives

\begin{equation}
\label{eq:intrinsics_constraints_aspect_ratio2}
\colvec{3}{u^{\prime}}{v^{\prime}}{f\left(1 + \sum_{n=1}^N \lambda_n
r_0(a)^{2n}\right)} \times \colvec{3}{x^{\prime}}{y^{\prime}}{\hat{\vh}_3\tr\vx}
= 0,
\end{equation}
where $r_0(a)=\sqrt{(\frac{u'}{a})^2+{(v')}^2}$,\, $x^{\prime}
= \hat\vh_1 \tr \vx$,\, $y^{\prime} = \hat\vh_2\tr \vx$. After
reparameterizing with $b = a^2$, it can be seen that for $N=1$,
\eqref{eq:intrinsics_constraints_aspect_ratio2} 
is linear in the unknowns and for $N=2$, a polynomial system of degree
three must be solved.

\begin{figure}[!t]
\begin{center}
\footnotesize
\includegraphics[height=0.28\columnwidth]{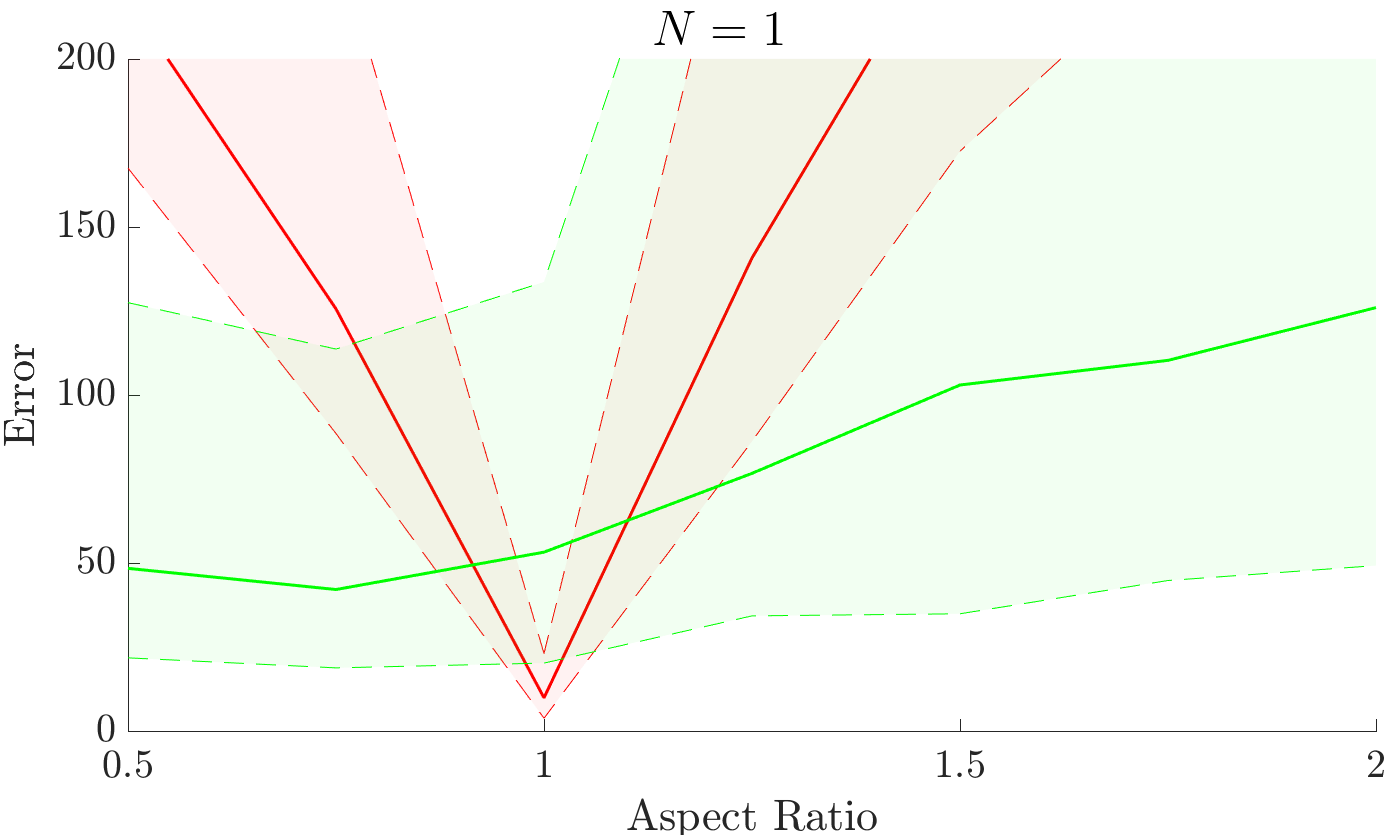}
\includegraphics[height=0.28\columnwidth]{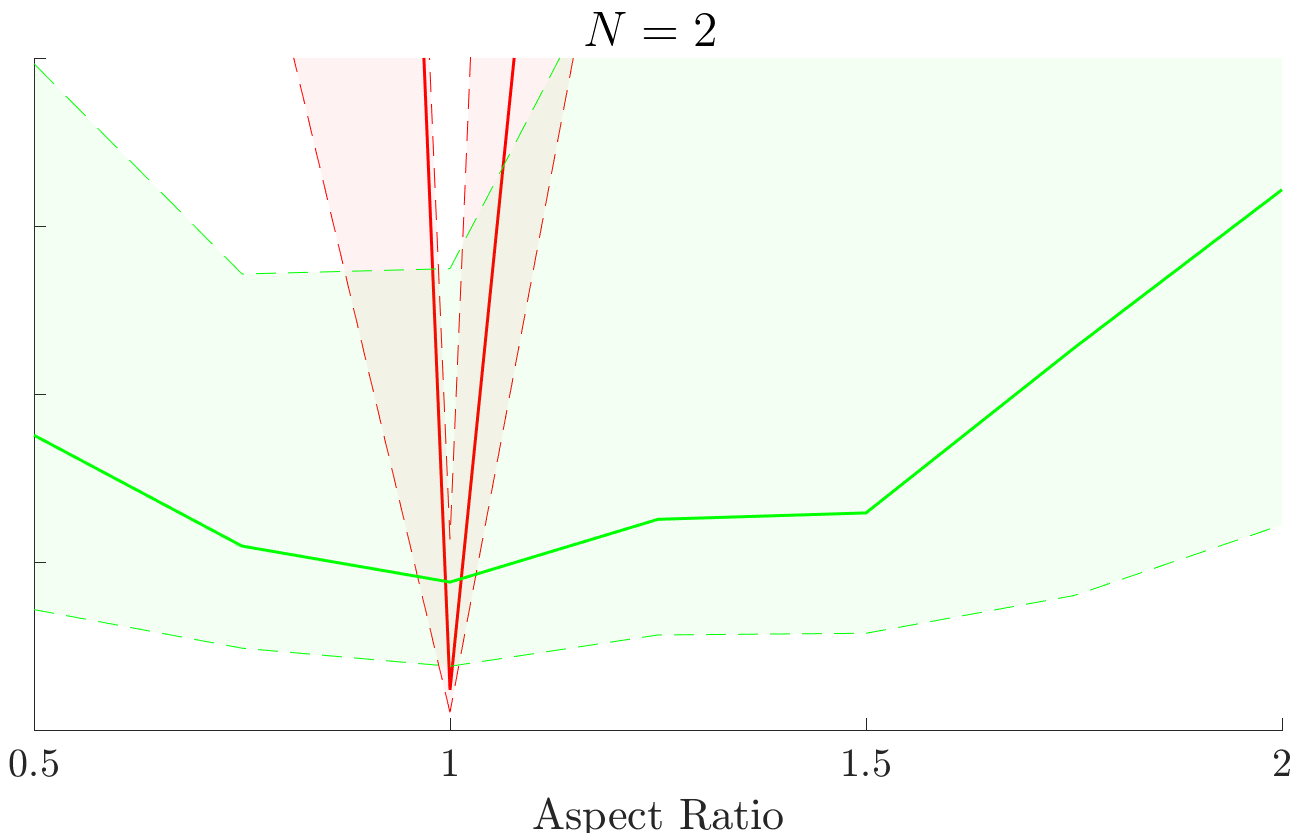}
\caption{\textbf{Noise sensitivity.}
RMS reprojection error on synthetic images 
with added 1 px noise for (left) the solvers for the division model
complexity $N=1$, and (right) the solvers for the complexity
$N=2$. Red curves correspond to the linear solvers
from \eqref{eq:intrinsic_solution} that do not solve for the aspect
ratio, and green curves correspond to the aspect ratio solvers
from \eqref{eq:intrinsics_constraints_aspect_ratio2}.  Solid curves
are the median errors and the shaded plots are the corresponding
interquartile ranges.}
\label{fig:supp_ar_solver}
\vspace{-22pt}
\end{center}
\end{figure}

We tested the solvers for the cases $N=1$ and $N=2$ on synthetic data
and found that they are sensitive to noise. With a sufficiently good
guess on aspect ratio, the solver in \eqref{eq:intrinsic_solution}
performs better than the aspect ratio solver
(see \figref{fig:supp_ar_solver}). We opted to sample over aspect
ratio rather than introduce these solvers. We leave the incorporation
of minimal solvers for unknown aspect ratio for future work.

\newpage
\quad
\vspace{45pt}

\section{Recovering Radial-Projection Functions for User-Selected Camera Models} 
\label{sec:supp_m2m_regression}
The detected corners $\vu$ can be back-projected to rays $\gamma g(\vu) = \gamma \rowvec{3}{u}{v}{\psi(r(\vu))}^{\T}, \gamma > 0$
in the direction of the board fiducials in the camera's coordinate
system. The polar angle of the ray determines how the points of the
ray are projected, so any point of the ray can be used (equivalently,
any $\gamma > 0$ in \eqref{eq:omni_camera_model} can be used).  The distances 
to the optical axis $R$
and principal plane $Z$ are computed from the unit vector concurrent
with the ray, $\frac{g(\vu)}{\|g(\vu)\|}$.
With the radii and depths recovered, least squares can be used
to recover the unknowns of the radial projection functions listed
in \tabsref{tab:models}. This demonstrates the ease with which
BabelCalib is extended.

\paragraph{Division Model to Kannala-Brandt Regression}
The experiments of \secref{sec:evaluation} confirm that the
Kannala-Brandt (KB) model is the most flexible and accurate over the
largest range of lenses, which is consistent with the results
of \cite{Usenko-3dv-2020}. We also found that the Kannala-Brandt model
is also effective for catadioptric rigs (see \tabref{tab:holdout-test}).

The number of failed calibrations of OpenCV and Kalibr reported
in \tabref{tab:holdout-test} suggests that Kannala-Brandt is one of 
the hardest camera models to initialize.
Directly computing a model proposal for Kannala-Brandt is
difficult. The displacement of the projection from the optical center
is proportional to a polynomial function of $\atantwo$.
The trigonometric function $\atantwo$ is not easily
eliminated since the unknown depth $Z$ is unique to each fiducial on
the chessboard. Thus the problem cannot be solved as a polynomial
system. BabelCalib initializes Kannala-Brandt by linearly regressing
its parameters against the estimated division model, which is
formulated in \secref{sec:good_guess}. We derive the linear system
here to show how easy the model-to-model regression is once the
back-projection function is known.

We back-project the corners $\vu_i
= \rowvec{3}{u_i}{v_i}{1} \tr$ to ray directions $\vxp[i]
= \rowvec{3}{x'_i}{y'_i}{z'_i} \tr$
where $\vxp[i] = g(\mK^{-1}\vu_i)$.
The radii and depths are
computed as $R_i=\sqrt{x'^2_i+y'^2_i}$ and $Z_i=z'_i$, respectively. The
polar angles $\omega_i = \atantwo{(R_i,Z_i)}$ are determined, and the unknown coefficients
of the polynomial $r_i=\omega_i+\sum^{4}_{n=1} k_n \omega_i^{2n+1}$
can be determined linearly,

\newpage
\begin{equation}
    \vspace{-20pt}
    \begin{bmatrix}
        & \vdots & \\
        \omega_i^3 & \ldots & \omega_i^9  \\
        \omega_{i+1}^3 & \ldots & \omega_{i+1}^9 \\        
        & \vdots & \\
    \end{bmatrix}
    \begin{pmatrix}
     k_1 \\ k_2 \\ k_3 \\ k_4
    \end{pmatrix} = 
    \begin{pmatrix}
       \vdots \\ r_i-\omega_i \\ r_{i+1}-\omega_{i+1} \\  \vdots 
    \end{pmatrix}.
    \label{eq:kb_intrinsic_solution}
    \vspace{20pt}
\end{equation}

Solving \eqref{eq:kb_intrinsic_solution} fully specifies the
Kannala-Brandt projection model since the center
of projection $\ve$ and the focal length $f$ are known from estimation
of the back-projection model --- from \eqref{eq:principal_point} and
\eqref{eq:intrinsic_solution}, respectively.
\vspace{20pt}

\section{Algorithm}
\label{sec:supp_algorithm}

To summarize the proposed RANSAC-based calibration
framework detailed in \secref{sec:robust_estimation}, we provide \algref{alg:method} with all the crucial steps of the BabelCalib framework. The details on the pose optimization are omitted in the algorithm.

\SetKwInput{KwData}{Input}
\SetKwInput{KwResult}{Output}
\SetKwInput{Parameter}{Parameters}
\begin{algorithm}[!h]
\small
\SetAlgoNoLine
\KwData{2D-3D point correspondences $\cspond{\vu_{ijk}}{\vX[ij]}$}
\Parameter{Radial projection model $\phi_\theta$}
\KwResult{$\Theta^\text{*}=\{\theta^\text{*},\mK^\text{*},\mR_{jk}^\text{*},\vt_{jk}^\text{*}\}$}
$\mathcal{J}_0^\text{*} \leftarrow \infty,\,\,
\mathcal{J}^\text{*} \leftarrow \infty$\\
\Repeat{$T$ \emph{iterations}}{
    Sample image $k'$ and plane $j'$\\
    Sample $\{\cspond{\vu_{ij'k'}}{\vX[ij']}\}_{i=1}^N$\\
    Compute $\ve$ and $\m F$ from $\{\cspond{\vu_{ij'k'}}{\vX[ij']}\}_{i=1}^N$ with \eqref{eq:geometric_correction_objective}\\
    Correct corners $\vu_{ij'k'}$ to $\vu^*_{ij'k'}$ with \eqref{eq:corner_correction}\\
    Compute $\{\mR_{j'k'}, t_x, t_y\}$ from  $\mF$ with \eqref{eq:orthonormality_constraints}\\
    Sample aspect ratio $a$ from $[0.5, 2]$\\
    Compute $\{f,\lambda_1,...,\lambda_N,t_z\}$ with \eqref{eq:intrinsic_solution}\\
    $\mK \leftarrow \mT(\ve)\diag{a f,f,1},\,\,
    \vt_{j'k'} \leftarrow \rowvec{3}{t_x}{t_y}{t_z}\tr$\\
    \If{$\phi^{-1}_\theta$ is not the division model}{
        Regress $\theta$ of $\phi_\theta$ with \eqref{eq:model_to_model_regression}\\
    }
    \For{$(k,j) \neq (k',j')$}{
        Sample image $k$ and plane $j$\\
        Back-project the corners $\vxp[ijk] \leftarrow g(\mK^{-1} \vu_{ijk})$\\
        Compute $\{\mR_{jk}$, $\vt_{jk}\}$ from $\{\vxp[ijk]\}_i$ with \cite{Nakano-BMVC-2019}\\
    }
    $\Theta_0 \leftarrow \{\theta,\mK,\mR_{jk},\vt_{jk}\}$\\
    Compute loss $\mathcal{J}_0 \leftarrow \mathcal{J}(\Theta_0)$ with \eqref{eq:robust_loss_pose}\\
    \If{$\mathcal{J}_0 < \mathcal{J}_0^*$}{
        $\mathcal{J}_0^* \leftarrow \mathcal{J}_0,\,\,
        \mathcal{I}_0^* \leftarrow \mathcal{I}_0$\\
        Optimize $\Theta_{LO} \leftarrow \operatorname{argmin}_{\Theta}\mathcal{J}(\Theta)$ with \eqref{eq:robust_loss_pose} \\
        Compute loss $\mathcal{J}_{LO} \leftarrow \mathcal{J}(\Theta_{LO})$ with \eqref{eq:robust_loss_pose} \\
        \If{$\mathcal{J}_{LO} < \mathcal{J}^*$}{
            $\mathcal{J}^* \leftarrow \mathcal{J}_{LO},\,\,
            \Theta^* \leftarrow \Theta_{LO}$\\
        }
        }
    }
    Compute inlier ratio $\mathcal{I}^* \leftarrow \mathcal{I}(\Theta^*)$ with \eqref{eq:inlier_ratio}\\
    \caption{BabelCalib Camera Calibration}
\label{alg:method}
\end{algorithm}

\begin{figure}[!b]
\begin{center}
\footnotesize
\begin{tabular*}{0.48\textwidth}{c@{\extracolsep{\fill}}c@{\extracolsep{\fill}}c@{\extracolsep{\fill}}c}
\texttt{GOPRO} & \texttt{ENTANIYA} & \texttt{TUMVI} \\
\includegraphics[height=0.097\textwidth]{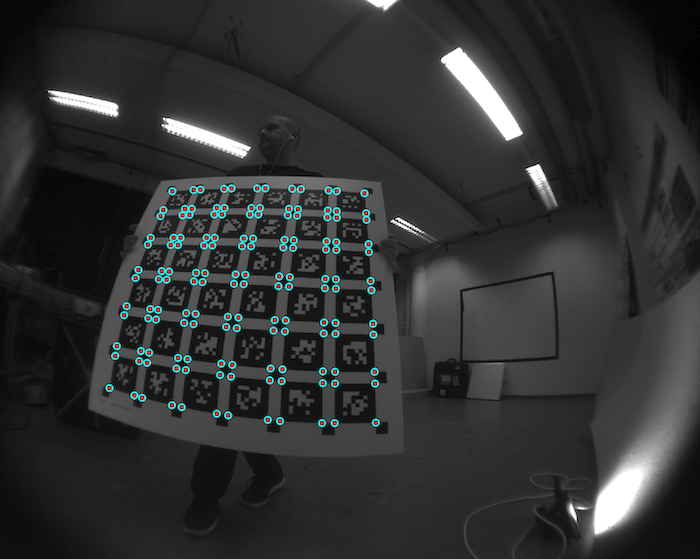} &
\includegraphics[height=0.097\textwidth]{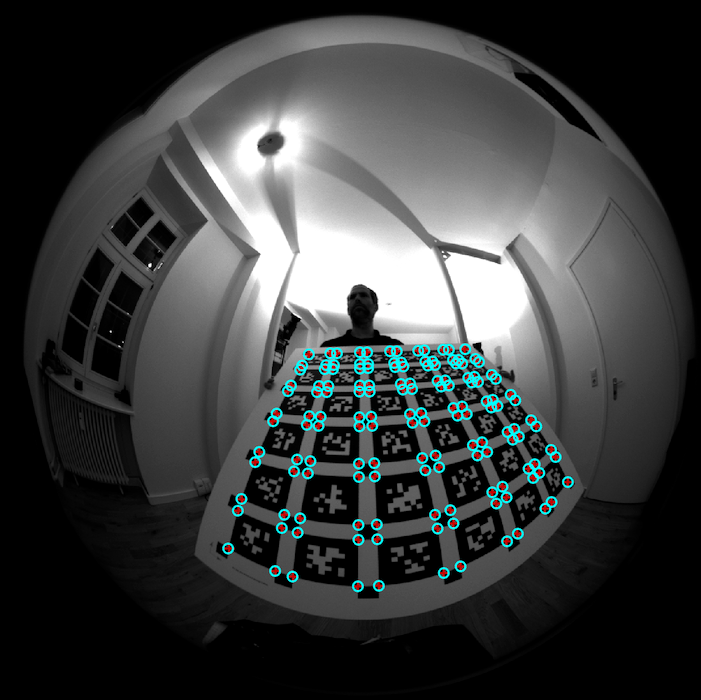} &
\includegraphics[height=0.097\textwidth]{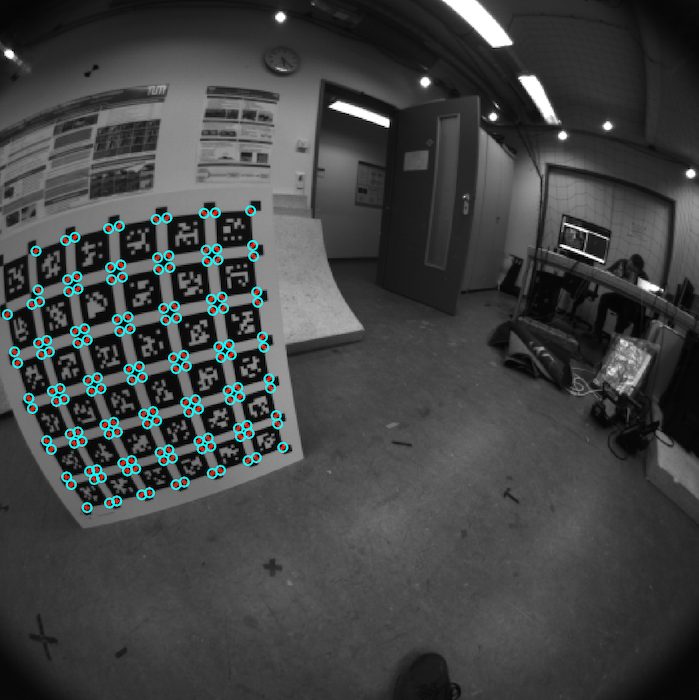}&
\\
\includegraphics[height=0.097\textwidth]{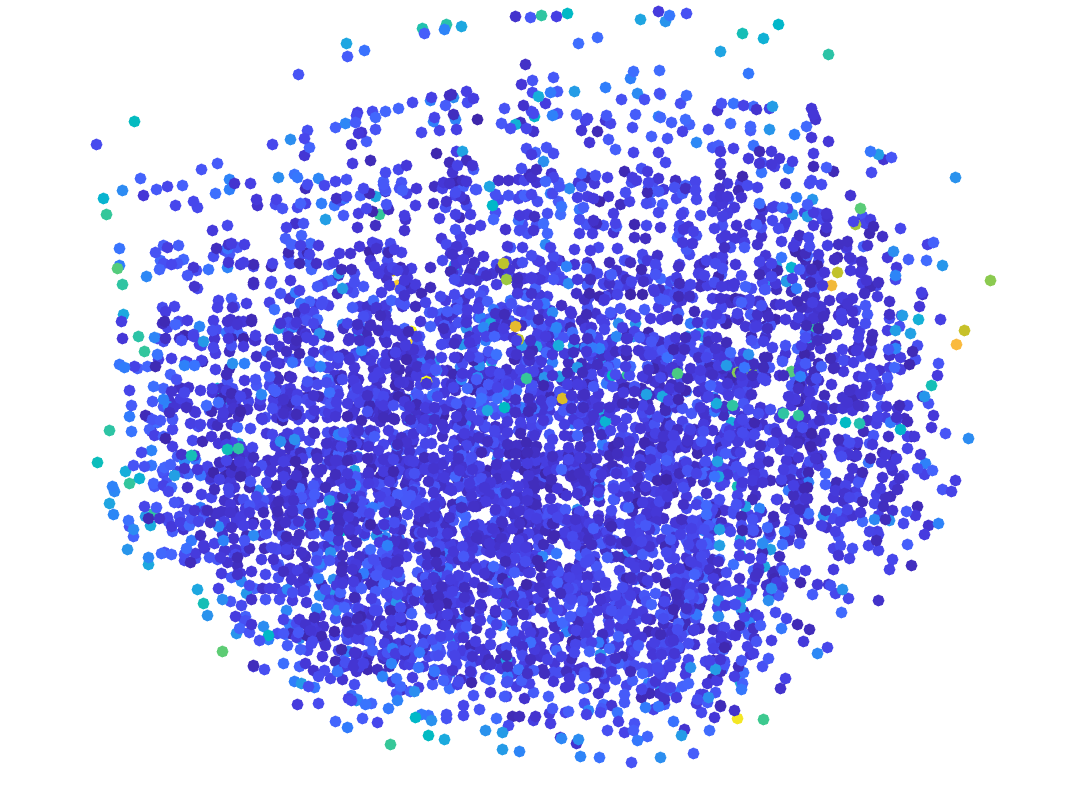} &
\includegraphics[height=0.097\textwidth]{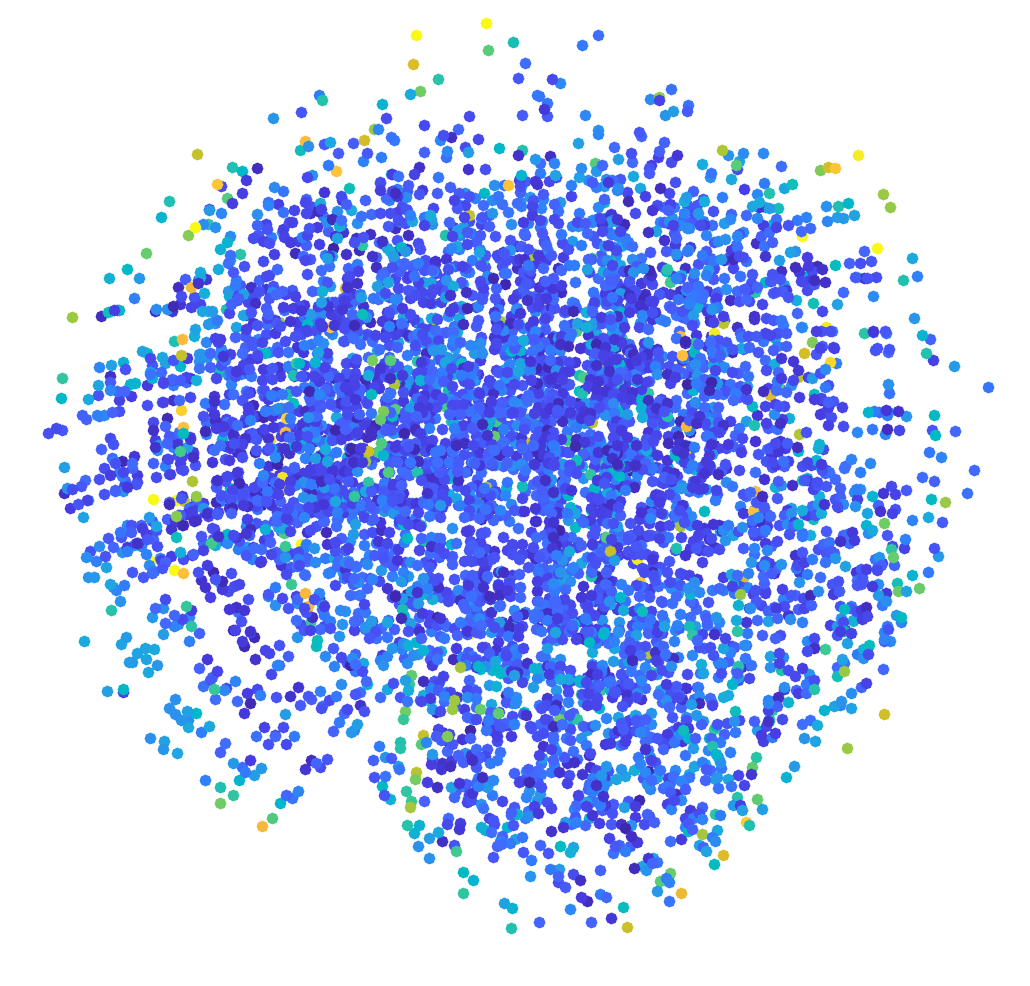} &
\includegraphics[height=0.097\textwidth]{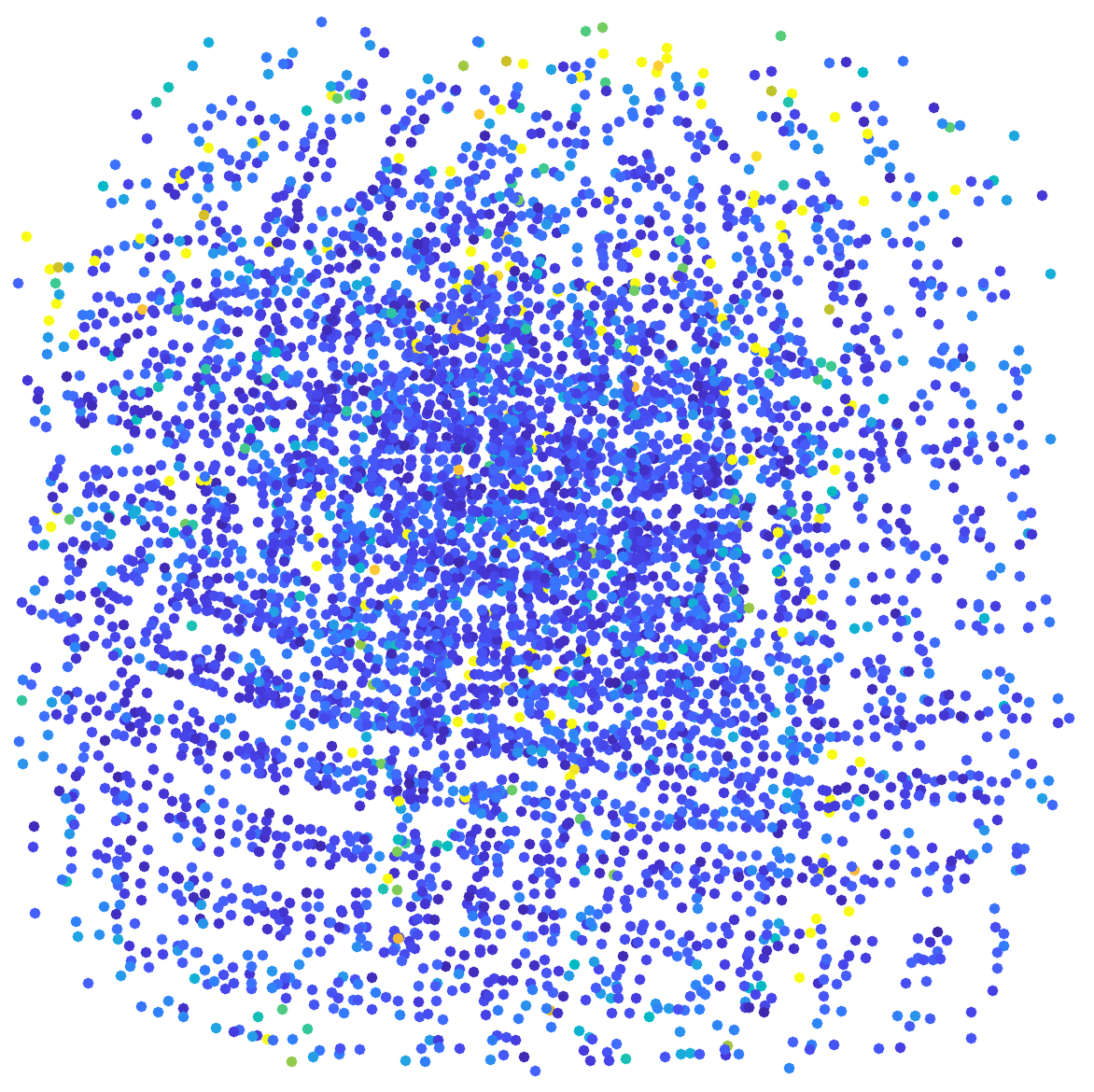}&\\
\texttt{KaidanOmni} & \texttt{MiniOmni} & \texttt{VMRImage}\\
\includegraphics[height=0.097\textwidth]{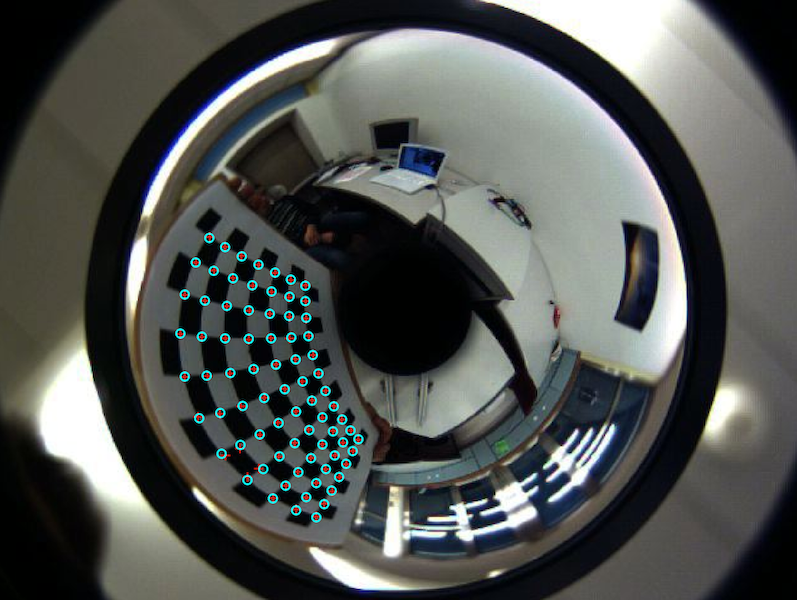}&
\includegraphics[height=0.097\textwidth]{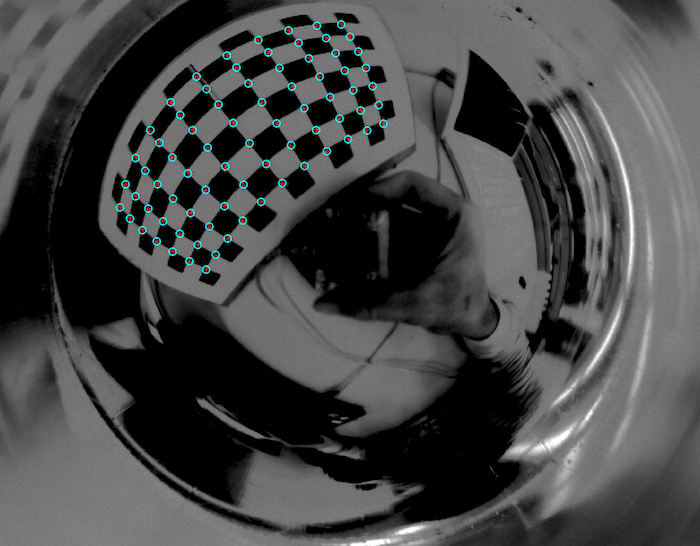} &
\includegraphics[height=0.097\textwidth]{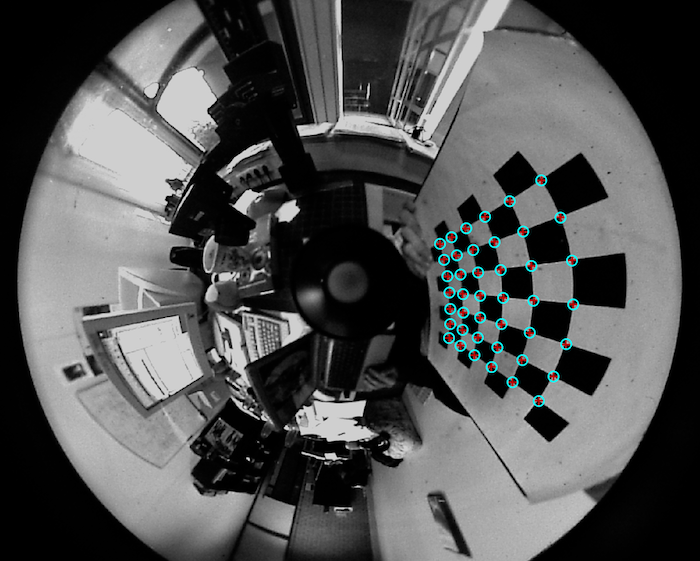}& 
\multirow{2}{*}[40pt]{\centering 
\includegraphics[height=0.194\textwidth]{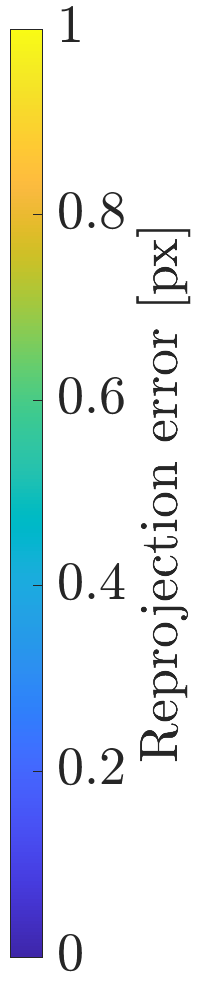}}\\
\includegraphics[height=0.097\textwidth]{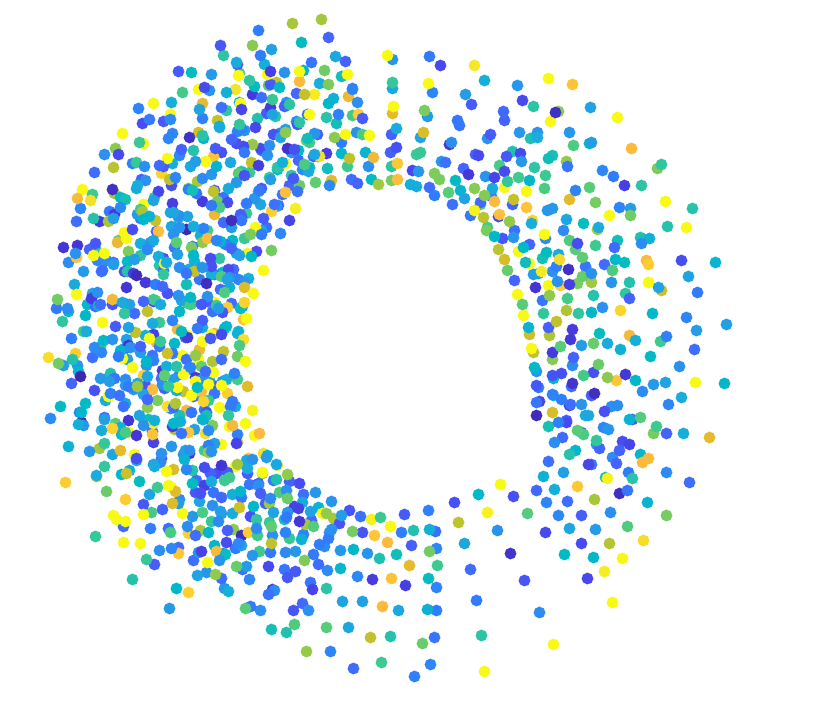} &
\includegraphics[height=0.097\textwidth]{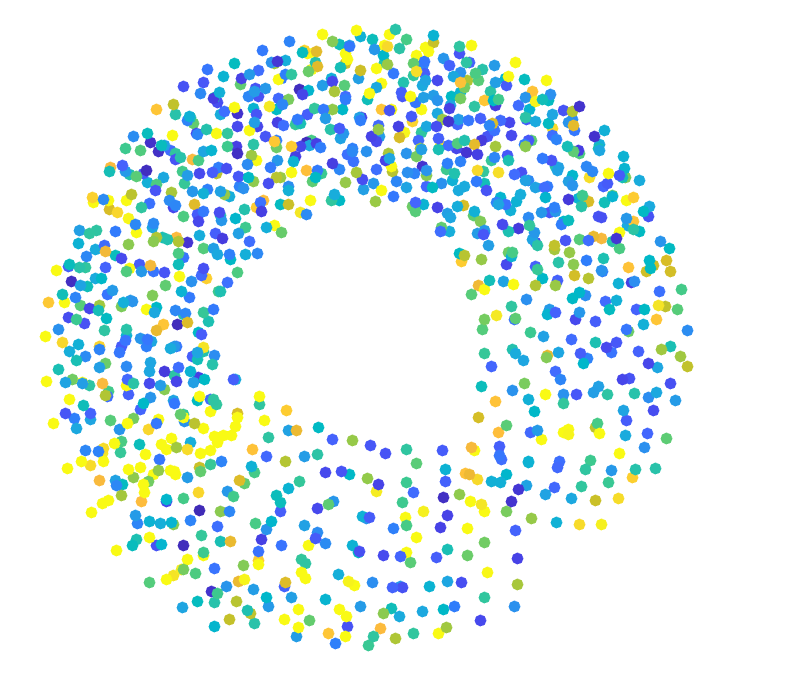} &
\includegraphics[height=0.097\textwidth]{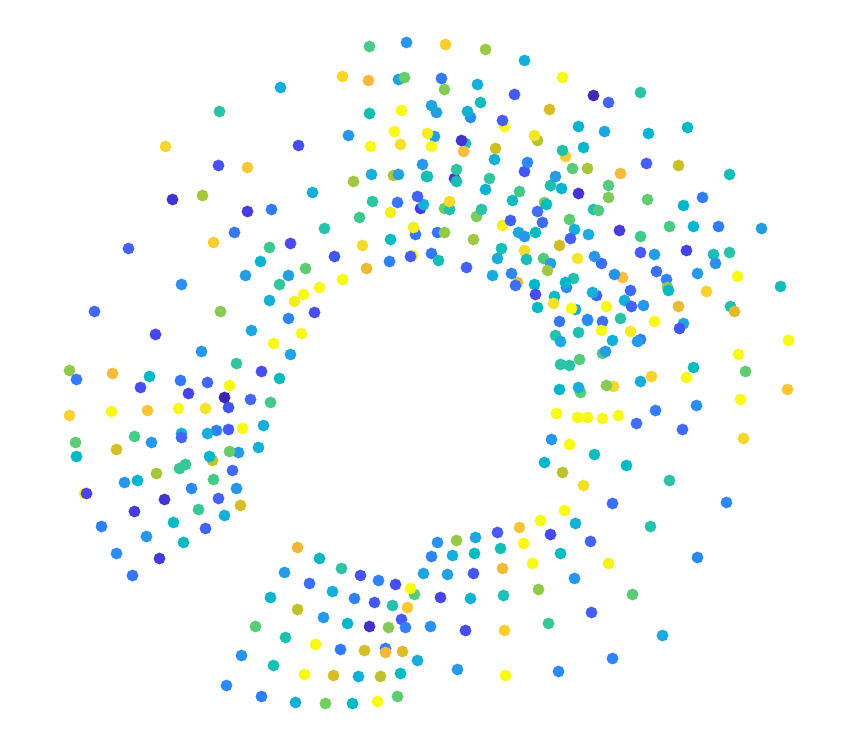}
& \\
\texttt{DAVIS-indoor}  & \texttt{DAVIS-outdoor} & \texttt{\footnotesize{Snapdragon}}\\
\includegraphics[height=0.097\textwidth]{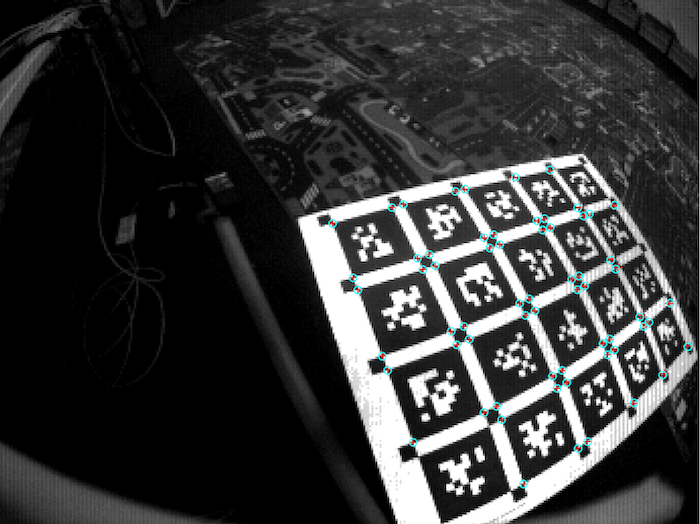} &
\includegraphics[height=0.097\textwidth]{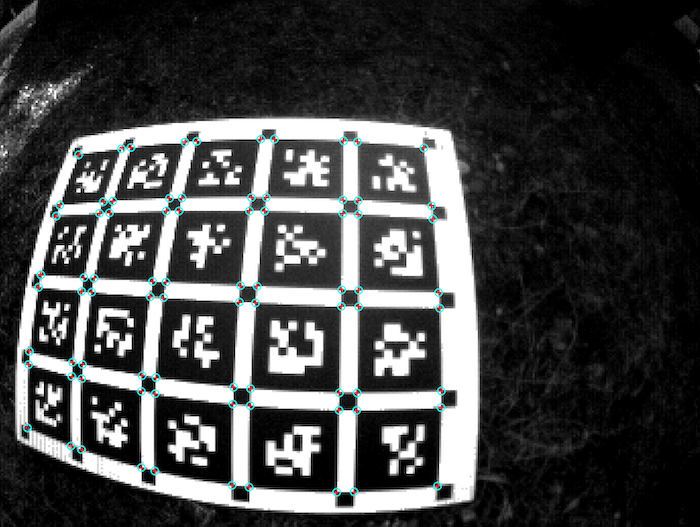} &
\includegraphics[height=0.097\textwidth]{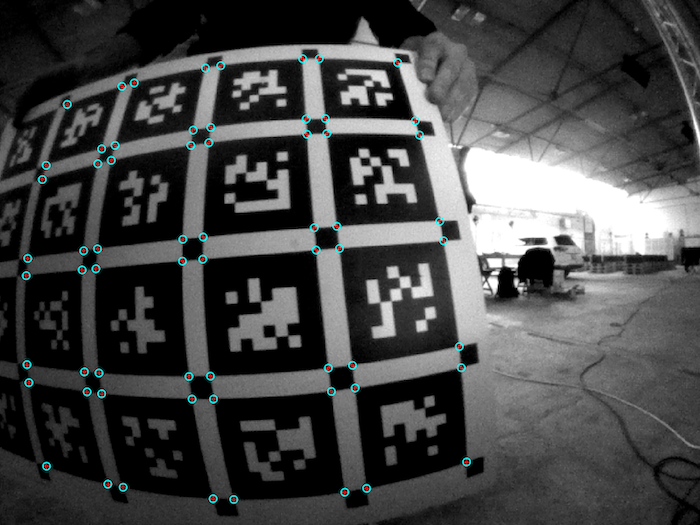}&
\\
\includegraphics[height=0.097\textwidth]{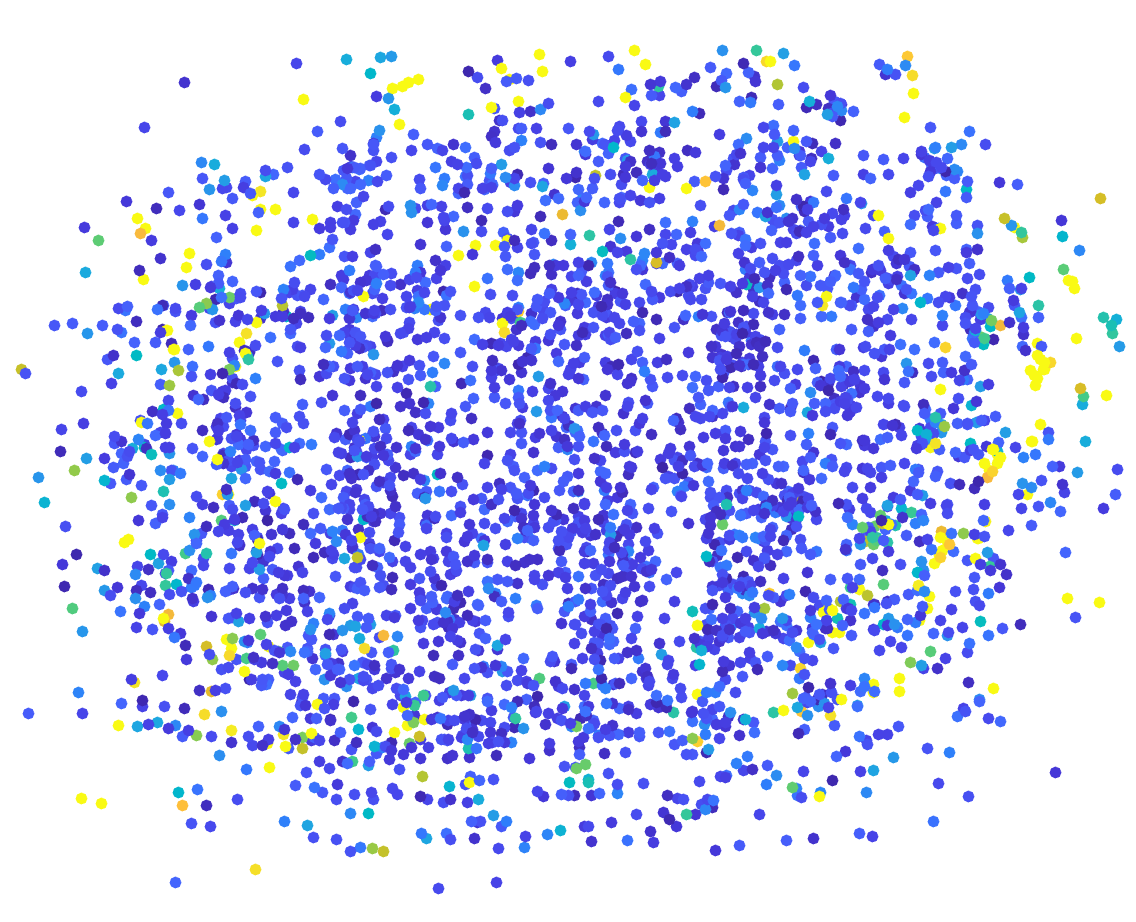} &
\includegraphics[height=0.097\textwidth]{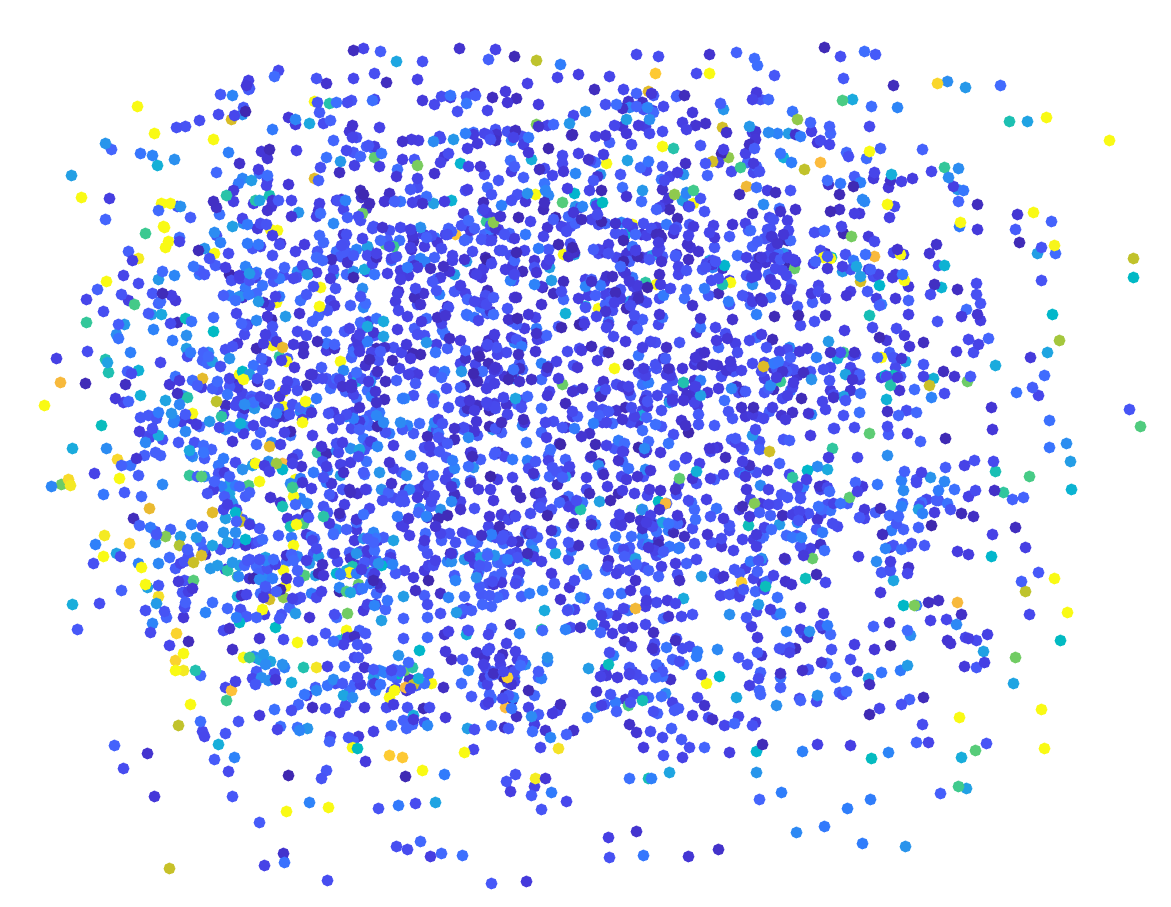} &
\includegraphics[height=0.097\textwidth]{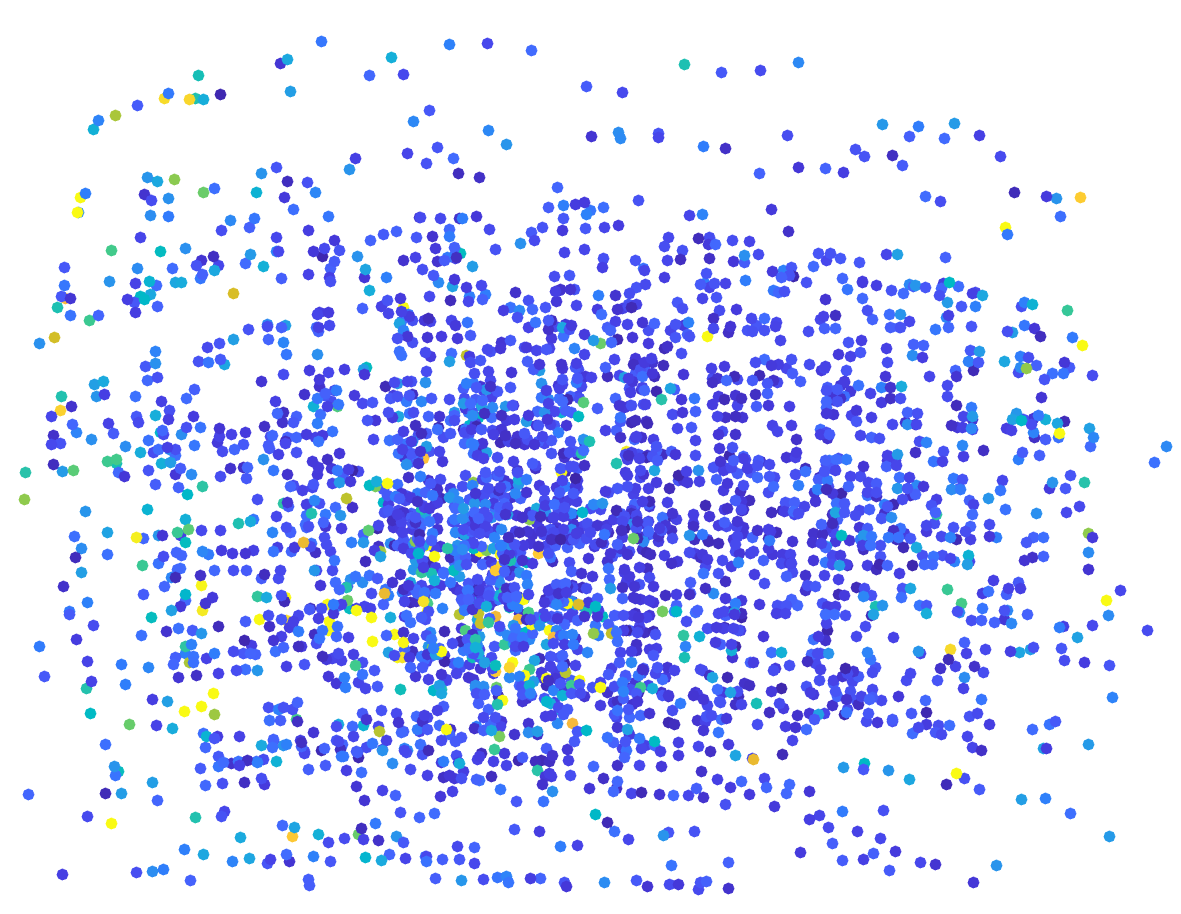}
& \\
\end{tabular*}
\caption{\textbf{Kalibr, OCamCalib and UZH data and calibration results.} 
(rows 1,3,5) example images with detected (red crosses) and reprojected (cyan circles) corners; (rows 2,4,6) all corners from the camera subset color-coded corresponding to their reprojection errors.}
\label{fig:supp_res}
\vspace{-15pt}
\end{center}
\end{figure}
\begin{figure*}[!t]
\begin{center}
\footnotesize
\begin{tabular}{c c c}
Train size: 1 & Train size: 2 & Train size: 5 \\
\includegraphics[height=0.16\textwidth]{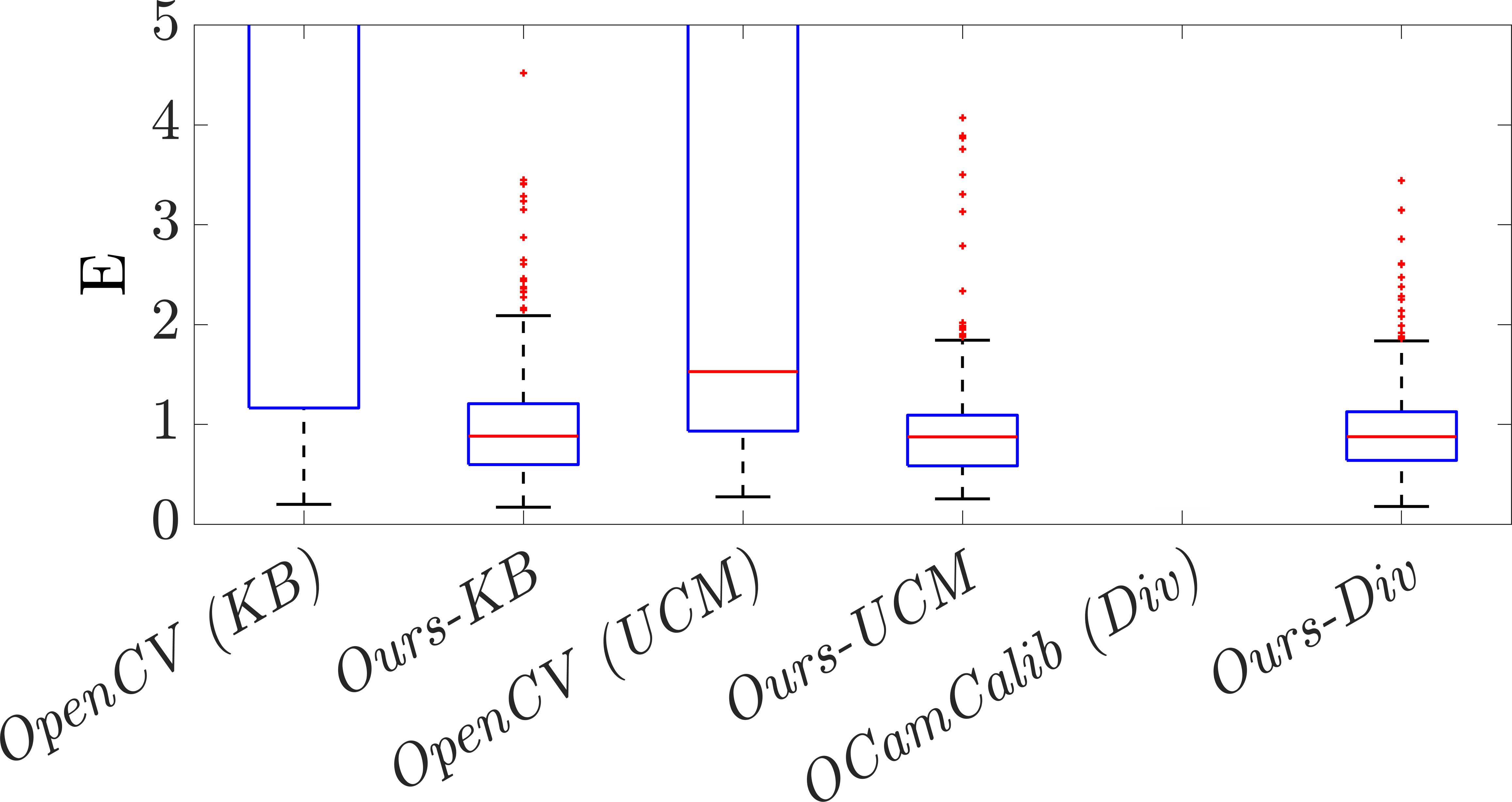} &
\includegraphics[height=0.16\textwidth]{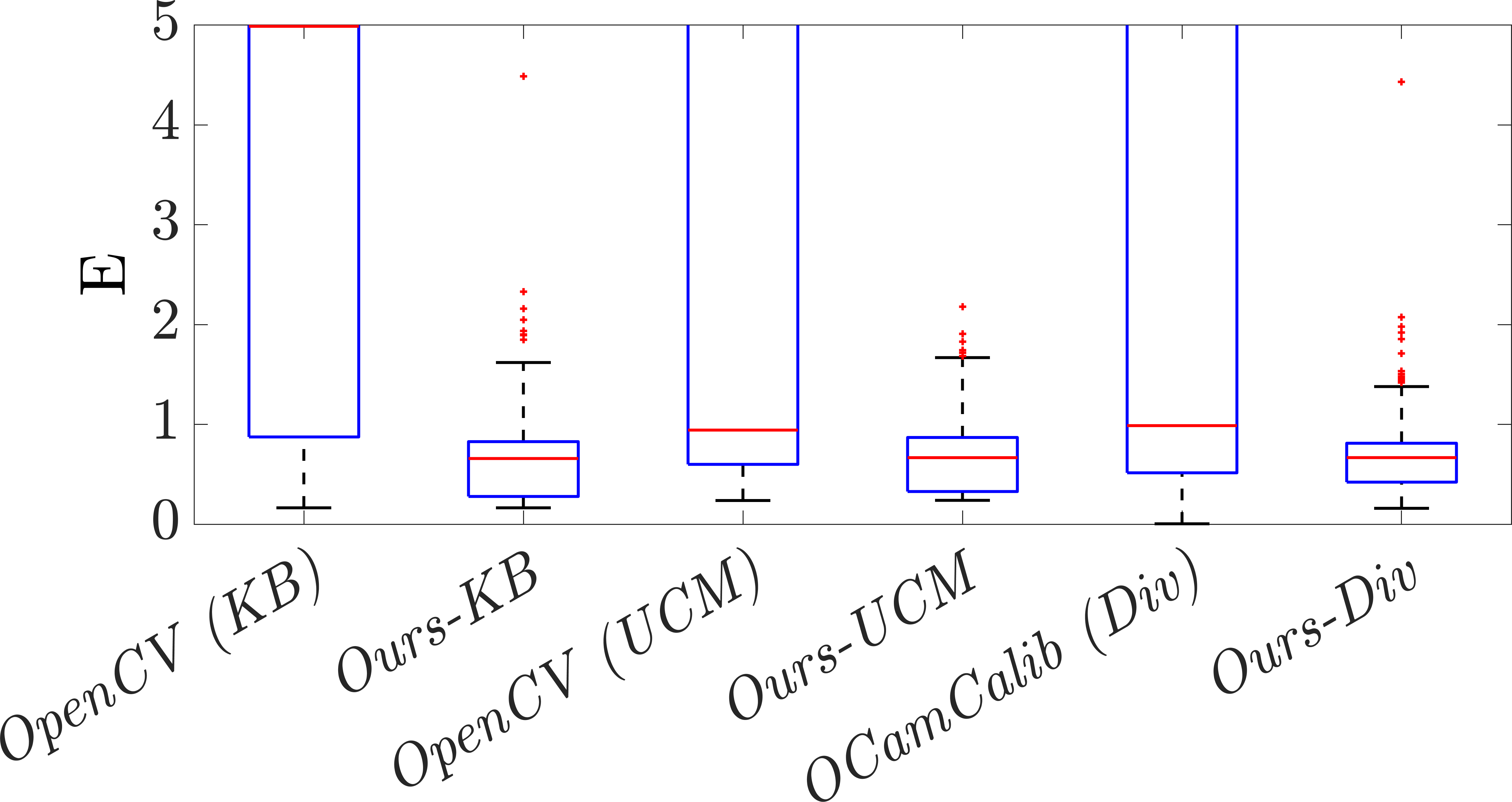} &
\includegraphics[height=0.16\textwidth]{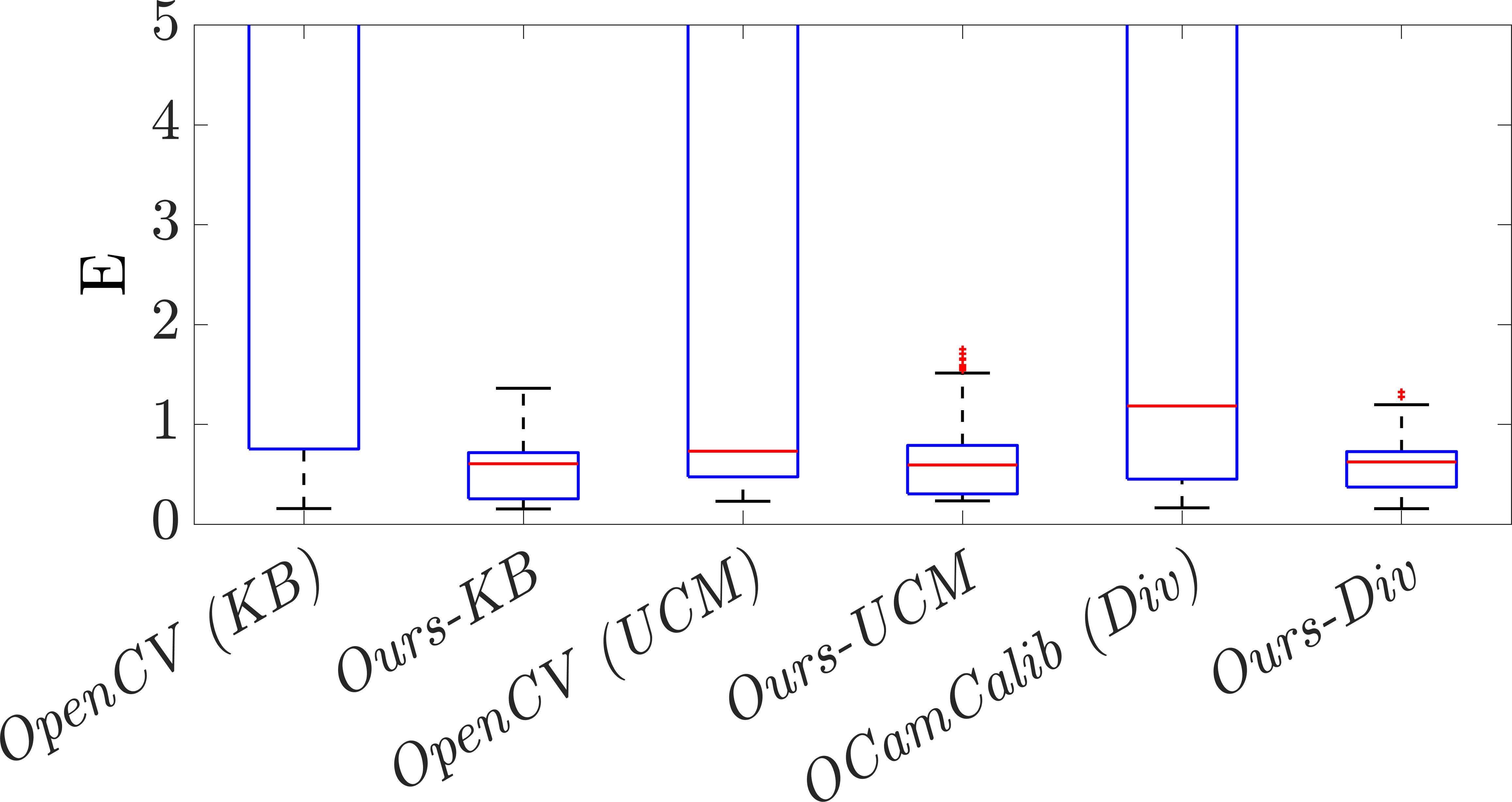} \\
\end{tabular}
\normalsize
\caption{\textbf{Calibration results for limited train data.} Cross-validation with corresponding train size is performed, and the reported errors are normalized in accordance with image resolution $1000\times1000 $ px. E denotes the RMS weighted reprojection error. \occDiv requires at least two images for calibration.}
\label{fig:crossval}
\vspace{-15pt}
\end{center}
\end{figure*}

\section{Calibration with Limited Data}
\label{sec:supp_limited_data}
Calibration accuracy is dependent on good feature coverage.
However, fast calibration from few images is also important for users e.g. for
lens inventory purposes.
We compared the performance of calibration methods on
a limited amount of training data starting from a single image. We randomly drew
the subsamples of sizes 1, 2, and 5 images from the training data (10 times for
each train size) and evaluated the calibrations on the hold-out test data. The
\figref{fig:crossval} reports the distributions of the robust test errors
\eqref{eq:robust_loss_pose}. To remove the effect of the image resolution for
different cameras, the errors are normalized in accordance with the resolution
$1000\times1000 $ px. There are several limitations found in current frameworks.
\occDiv requires at least two images for calibration so they don't have the
results for the train size 1. Also, as was mentioned in previous section, this
toolbox requires all points to be visible from all views which is a big
limitation. All other methods would occasionally fail for particular subsets of
images. For such failure cases we set the test error to be the maximum error.
The proposed calibration method never fails and provides the best calibration
starting from a single image already.